%% file: main.tex
\documentclass[sigconf, screen]{acmart}

\AtBeginDocument{%
  }

\setcopyright{acmlicensed}
\copyrightyear{2024}
\acmYear{2024}
\acmDOI{XXXXXXX.XXXXXXX}

\acmConference[Conference acronym 'XX]{Make sure to enter the correct
  conference title from your rights confirmation emai}{June 03--05,
  2018}{Woodstock, NY}
\acmISBN{978-1-4503-XXXX-X/18/06}

\usepackage{graphicx}
\usepackage{amsmath}
\usepackage{amsfonts}
\usepackage{multirow}
\usepackage{booktabs}
\usepackage{subfigure}
\usepackage{amsthm}
\usepackage{mathrsfs}
\usepackage{stfloats}
\usepackage{multicol}

\usepackage{array} 
\newcolumntype{C}[1]{>{\centering\arraybackslash}p{#1}}

\usepackage[many]{tcolorbox}

\NewTColorBox[auto counter]{NewBox}{ s O{!htbp} m m }{%
  floatplacement={#2},                        
  IfBooleanTF={#1}{float*,width=\textwidth}{float},  
  fontupper=\small, 
  title={Prompt \thetcbcounter: #4},
  label={#3}                                  
}

\begin{document}

\title{See it, Think it, Sorted: Large Multimodal Models are Few-shot Time Series Anomaly Analyzers}

\author{Jiaxin Zhuang}
\email{zhuangjx23@mails.tsinghua.edu.cn}
\affiliation{%
  \institution{Tsinghua University}
  \city{Beijing}
  \country{China}
}

\author{Leon Yan}
\email{yansc23@mails.tsinghua.edu.cn}
\affiliation{%
  \institution{Tsinghua University}
  \city{Beijing}
  \country{China}
}

\author{Zhenwei Zhang}
\authornote{Project Lead}
\email{zzw20@mails.tsinghua.edu.cn}
\affiliation{%
  \institution{Tsinghua University}
  \city{Beijing}
  \country{China}
}

\author{Ruiqi Wang}
\email{wang-rq23@mails.tsinghua.edu.cn}
\affiliation{%
  \institution{Tsinghua University}
  \city{Beijing}
  \country{China}
}

\author{Jiawei Zhang}
\email{jiawei-z23@mails.tsinghua.edu.cn}
\affiliation{%
  \institution{Tsinghua University}
  \city{Beijing}
  \country{China}
}

\author{Yuantao Gu}
\authornote{Corresponding Author}
\email{gyt@tsinghua.edu.cn}
\affiliation{%
  \institution{Tsinghua University}
  \city{Beijing}
  \country{China}
}


\begin{abstract}

Time series anomaly detection (TSAD) is becoming increasingly vital due to the rapid growth of time series data across various sectors. Anomalies in web service data, for example, can signal critical incidents such as system failures or server malfunctions, necessitating timely detection and response. However, most existing TSAD methodologies rely heavily on manual feature engineering or require extensive labeled training data, while also offering limited interpretability.
To address these challenges, we introduce a pioneering framework called the Time series Anomaly Multimodal Analyzer (\textbf{TAMA}), which leverages the power of Large Multimodal Models (LMMs) to enhance both the detection and interpretation of anomalies in time series data. By converting time series into visual formats that LMMs can efficiently process, TAMA leverages few-shot in-context learning capabilities to reduce dependence on extensive labeled datasets.
Our methodology is validated through rigorous experimentation on multiple real-world datasets, where TAMA consistently outperforms state-of-the-art methods in TSAD tasks. Additionally, TAMA provides rich, natural language-based semantic analysis, offering deeper insights into the nature of detected anomalies. Furthermore, we contribute one of the first open-source datasets that includes anomaly detection labels, anomaly type labels, and contextual descriptions—facilitating broader exploration and advancement within this critical field.
Ultimately, TAMA not only excels in anomaly detection but also provide a comprehensive approach for understanding the underlying causes of anomalies, pushing TSAD forward through innovative methodologies and insights.
\end{abstract}




\maketitle

\section{Introduction}
\label{sec:intro}
Web services have undergone significant expansion and advancement in recent years \cite{bouasker_qos_2020, yuan_energy-efficient_2022}. This expansion has led to the generation of vast amounts of time series data, including key performance indicators (KPIs) from cloud center and wireless base stations \cite{xu_unsupervised_2018, yu_supervised_2024}. Anomalies—defined as unexpected deviations from typical patterns in this data—can signal critical events such as device malfunctions and system failures. Consequently, time series anomaly detection (TSAD) techniques for web services have attracted considerable attention \cite{chen_lara_2024, nam_breaking_2024}, demonstrating significant practical value in monitoring web systems and ensuring service quality. Despite the advancements in TSAD methodologies, existing approaches often struggle with several key challenges. 


\begin{figure}[t]
    \setlength{\abovecaptionskip}{0pt}
    \setlength{\belowcaptionskip}{0pt}
    \centering
    \includegraphics[width=0.95\columnwidth]{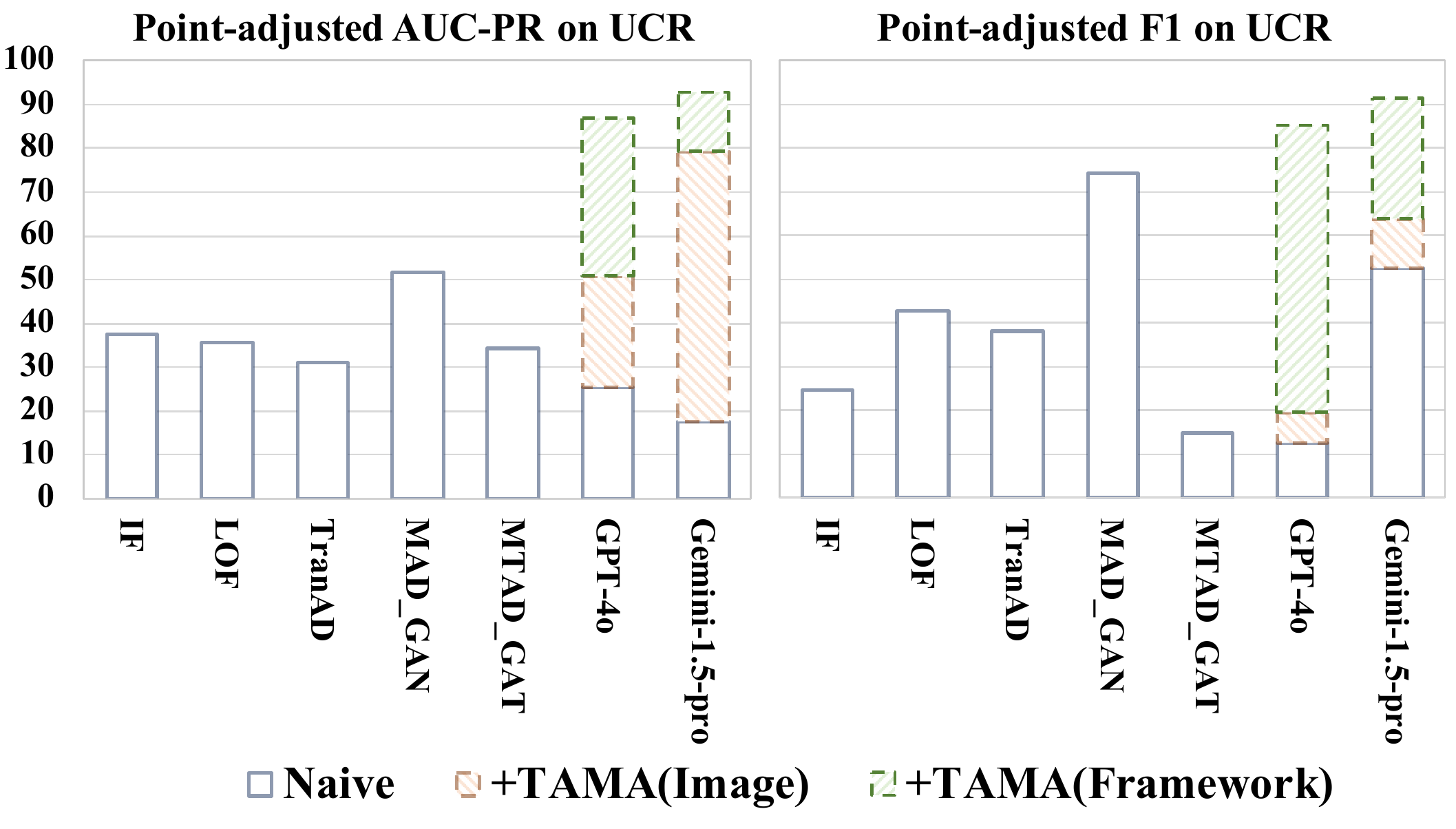}
    \caption{Comparison of {\emph{AUC-PR}} and \emph{F1} with PA on the UCR dataset. Models include the machine learning (IF, LOF), the deep learning (TranAD, MAD\_GAN, MTAD\_GAT), and LMMs ~(GPT-4o, Gemini-1.5-pro)}
    \label{fig:Intro}
    \vspace{-20pt}
\end{figure}

Firstly, different methods specialize in different datasets \cite{zamanzadeh_darban_deep_2024,app13031778}, and an "one-size-fits-all" universal solution is missing in the filed of TSAD \cite{schmidl2022anomaly}. Diverse approaches in TSAD demonstrate ongoing advancements, each contributing unique methodologies to tackle the inherent challenges of this field. Classical machine learning (ML) methods  \cite{liu_isolation_2008,feng_systematic_2019, ramaswamy_efficient_2000, yairi_fault_2001, chen_xgboost_2016, huang_lof-based_2013} are frequently based on strong assumptions or require empirically crafted manual features \cite{usmani_review_2022}. In contrast, deep learning ~(DL) techniques are heavily reliant on hyper-parameters and are either supervised or semi-supervised \cite{10043819, chalapathy_deep_2019, pang_deep_2022}, necessitating normal data for training--- except for \cite{audibert_usad_2020,zong_deep_2018,xu_unsupervised_2018} where other forms of training are required. Noticing that for most real-world datasets, assumptions needed by ML techniques usually do not hold, and training data without anomaly for DL methods are often undesirable, rare, or even unavailable. 

Secondly, most existing techniques offer insufficient interpretability, providing a limited understanding of the reasons behind how the anomalies are identified \cite{jacob_exathlon_2021}. Recently, more works have discussed and attempted to improve the explainability of TSAD algorithms \cite{jacob_exathlon_2021, lee_explainable_2023}, which is also considered a critical disadvantage of most deep learning methods. Toward this topic, the taxonomy of anomalies has been often mentioned in previous works \cite{blazquez-garcia_review_2021,choi_deep_2021}, where anomalies are categorized into two general types~(point and pattern) and several fertilized types \cite{lai2021revisiting}. However, due to the lack of high quality classification labels, few studies have achieved the semantic classification of anomalous data \cite{kiersztyn_detection_2022, pang_deep_2022}. 

Large language models ~(LLMs) have revolutionized natural language processing by demonstrating extraordinary capabilities in generalizing across numerous tasks \cite{naveed_comprehensive_2023,min_recent_2023}. LLMs can function as few-shot or even zero-shot learners \cite{brown_language_2020, gruver_large_2024}, which effectively mitigates the challenge of limited availability of anomaly-free training datasets in TSAD.
While LLMs are suitable for solving general problems, there have been attempts to apply them to time series analysis \cite{jin_time-llm_2023,su_large_2024,gruver_large_2024}, particularly for forecasting tasks, by trivially feed time series data into LLMs as text \cite{liu_lstprompt_2024}. However, studies have shown that LLMs without fine-tuning fail to achieve comparable results to existing task-specific approaches in time series tasks \cite{merrill_language_2024}, especially TSAD \cite{elhafsi_semantic_2023,alnegheimish_large_2024}. One interpretation for this is that LLMs are trained on text data and are insensitive to numerical values \cite{qian_limitations_2022,ye_towards_2024}, making them inherently unsuitable for capturing anomalies with small changes in amplitude \cite{choi_deep_2021}. 

To overcome the limitations of LLMs in handling numerical time series data, we introduce an innovative approach: first converting time series into images (``see it''), then utilizing Large Multimodal Models (LMMs) to analyze the visualized time series (``think it''), and finally detecting anomalous intervals with detailed explanations (``sorted''). Recent advancements reveal that LMMs can serve as multimodal reasoners, capable of integrating and analyzing diverse data types such as text, images, and audio \cite{wang_exploring_2024,zhang_mm-llms_2024}. This allows LMMs to perform complex tasks, including abstraction of images (e.g., tables and charts) \cite{zhang2024multimodalselfinstructsyntheticabstract}. With these capabilities, LMMs can effectively analyze visualized time series.

Building upon these advancements, we present a novel LMM-based framework named \textbf{T}ime series \textbf{A}nomaly \textbf{M}ultimodal \textbf{A}nalyzer (\textbf{TAMA}). TAMA goes beyond traditional TSAD approaches by not only identifying anomalies but also providing comprehensive anomaly type classification and supporting its decisions with detailed reasoning. This framework converts time series into chart inputs for LMMs, enabling effective analysis and interpretability. Our proposed anomaly analysis pipeline consists of three stages: Multimodal Reference Learning, Multimodal Analyzing, and Multi-scaled Self-reflection. This innovative framework enhances the analytical capabilities of LMMs, paving the way for more efficient, generalized, and interpretable anomaly detection in time series.

Our contributions are summarized as follows: (i) \textbf{A novel framework hybrid with the LMM}: we introduce a powerful LMM-based framework with an innovative series analysis pipeline designed for TSAD, while providing robust descriptions and semantic classifications. (ii) \textbf{An open-sourced dataset}: we have constructed and open-sourced one of the first dataset that includes anomaly detection labels, classification labels, and contextual descriptions associated with time series data. (iii) \textbf{Superior performance}: Extensive experiments across multiple TSAD datasets from various domains demonstrate that our proposed framework outperforms state-of-the-art methods.

\section{Relate Works}

\subsection{Time Series Anomaly Detection.} 
Many surveys \cite{chalapathy_deep_2019,pang_deep_2022,blazquez-garcia_review_2021, choi_deep_2021} are available in the field of TSAD. Classical methods \cite{ramaswamy_efficient_2000,yairi_fault_2001,chen_xgboost_2016}, especially unsupervised methods such as Isolation Forest ~(IF) \cite{liu_isolation_2008, bandaragoda_efficient_2014}, and Local Outlier Factor ~(LoF) \cite{huang_lof-based_2013} are introduced into TSAD in early stages. There are also variants of these classical ML algorithms like Deep Isolation Forest ~(DIF) \cite{xu_deep_2023}, which enhances IF by introducing non-linear partitioning. ML methods methods perform exceptionally well on many TASD datasets \cite{wu2021current,app13031778}, have been applied widely in industry \cite{usmani_review_2022}, and serve as strong baselines in recent researches.
 Deep learning methods focus on learning a comprehensive representation of the entire time series by reconstructing the original input or forecasting using latent variables. Among all reconstructing-based models, MAD-GAN \cite{li_mad-gan_2019} is an LSTM-based network enhanced by adversarial training. Similarly, USAD \cite{audibert_usad_2020} is an autoencoder-based framework that also utilizes adversarial training. MSCRED \cite{zhang_deep_2018} is designed to capture complex inter-modal correlations and temporal information within multivariate time series. 
 OmniAnomaly \cite{su_robust_2019} addresses multivariate time series by using stochastic recurrent neural networks to model normal patterns, providing robustness against variability in the data. MTAD-GAT \cite{zhao_multivariate_2020} employs a graph-attention network based on GRU to model both feature and temporal correlations. TranAD \cite{tuli2022tranad}, a transformer-based model, utilizes an encoder-decoder architecture that facilitates rapid training and high detection performance. Except reconstructing-based method, GDN \cite{deng_graph_2021} is a forecasting-based model that utilizes attention-based forecasting and deviation scoring to output anomaly scores. Additionally, LARA \cite{chen_lara_2024}, is a light-weight approach based on deep variational auto-encoders.The novel ruminate block and retraining process makes LARA exceptionally suitable for online applications like web services monitoring.



The aforementioned approaches have their strengths and weaknesses, with every model excelling in specific types of datasets while also exhibiting limitations. For instance, the ML techniques have been foundational, but they often require extensive feature engineering and struggle with complex datasets \cite{chalapathy_deep_2019}. For DL approaches,
reconstruction or forecasting-based models rely on reconstruction
error to identify anomalies, they are more sensitive to large amplitude anomalies and may fail to detect subtle pattern differences or
anomalies with small amplitude \cite{lee_explainable_2023}. In contrast, our proposed method can effectively capture anomalies with slight fluctuations by converting time series into images, and archive accurate few-shot detection result exploiting LMMs' splendid generalization ability.

\subsection{Time Series Anomaly Analysis.} 

Through a review of existing literature, we found that there is a lack of analysis on anomalies in current research. Common methods for analyzing anomalies identified by models involve visualizing the learned anomaly scores or parameters in relation to the ground truth \cite{dai_graph-augmented_2022, lee_explainable_2023}, as well as taxonomy of the anomalies \cite{blazquez-garcia_review_2021,choi_deep_2021, fahim_anomaly_2019}. Yet, limited research has investigated the efficacy of proposed models in classifying different types of anomalies. For instance, \cite{leon-lopez_anomaly_2022} introduced a framework based on Hidden Markov Models for anomaly detection, supplemented by an additional supervised classifier to identify potential anomaly types. GIN \cite{wang_anomaly_2024} employs a two-stage algorithm that first detects anomalies using an informer-based framework enhanced with graph attention embedding, followed by classification of the detected anomalies through prototypical networks.
Both aforementioned models rely on supervised training for their anomaly classification processes; consequently, the corresponding experiments conducted in these studies are limited to single classification datasets. In contrast, leveraging the capabilities of LLMs allows for not only the identification of anomalous data points but also the provision of specific classifications and potential underlying causes for these anomalies, articulated in natural language and achieved in an unsupervised manner.


\subsection{LLMs for time series.} 
Being pre-trained on enormous amounts of data, LLMs hold general knowledge that can be applied to numerous downstream tasks \cite{naveed_comprehensive_2023,min_recent_2023,chang_survey_2024}. Many researchers attempted to leverage the powerful generalization capabilities of LLMs to address challenges in time series tasks \cite{jin_time-llm_2023,su_large_2024, li_urbangpt_2024,elhafsi_semantic_2023}. For instance, Gruver et al. \cite{gruver_large_2024} developed a time series pre-processing scheme designed to align more effectively with the tokenizer used by LLMs. This approach can be illustrated as follows:
\begin{equation*}
0.123, 1.23, 12.3, 123.0 \rightarrow \text{" 1 2 , 1 2 3 , 1 2 3 0 , 1 2 3 0 0"}.
\end{equation*}
Additionally, LSTPrompt \cite{liu_lstprompt_2024} customizes prompts specifically for short-term and long-term forecasting tasks. Meanwhile, Time-LLM \cite{jin_time-llm_2023} reprograms input time series data using text prototypes and introduces the Prompt-as-Prefix (PaP) technique to further enhance the integration of textual and numerical information. Similarly, SIGLLM \cite{alnegheimish_large_2024} is an LLM-based framework for anomaly detection with a moudle to convert time series data into language modality. Most of these efforts focus primarily on forecasting tasks and are largely confined to textual modalities. 


Existing works remain constrained by the limited availability of sequential samples in the training datasets of LLMs \cite{merrill_language_2024} and the models' inherent insensitivity to numerical data \cite{qian_limitations_2022,ye_towards_2024}. Consequently, LLMs struggle to capture subtle changes in time series, making it difficult to produce reliable results \cite{merrill_language_2024}. Consequently, while we recognize that natural language is a modality in which LLMs excel, it may not be the most effective format for processing time series data.

\subsection{Time series as images}
The concept of transforming time series data into images has gained significant attention in recent years. One prominent method \cite{cao_image_2021} highlights the effectiveness of this approach, demonstrating its ability to improve recognition rates by utilizing hierarchical feature representations from raw data. Another innovative work \cite{wang_encoding_2014} introduced Gramian Angular Fields (GAF) and Markov Transition Fields (MTF) as methods to encode time series data into images, and this technique is further explored in \cite{wang_imaging_2015}. \cite{li_time_2023} presents a novel perspective by converting irregularly sampled time series into line graph images, and utilizing pre-trained vision transformers for classification. Additionally, TimesNet \cite{wu_timesnet_2023} is a time series analysis foundation model that exploit CV advancing techniques by converting time-series to 2D tensors. 

With the emergence of LMMs \cite{zhang_mm-llms_2024}, there is potential for enhanced reasoning capabilities that can accommodate a broader range of tasks beyond single-modal textual inputs \cite{wang_exploring_2024,zhang_multimodal_2024}. Some research has indicated that these models possess analytical abilities for interpreting charts \cite{zhang2024multimodalselfinstructsyntheticabstract}; however, no studies have yet applied them to the domain of anomaly detection in time series data. This gap highlights the need for further exploration into how LMMs can be effectively utilized to detect and analyze anomalies based on visualized time series data.

\begin{figure*}[t]
\centering
\includegraphics[width=\linewidth]{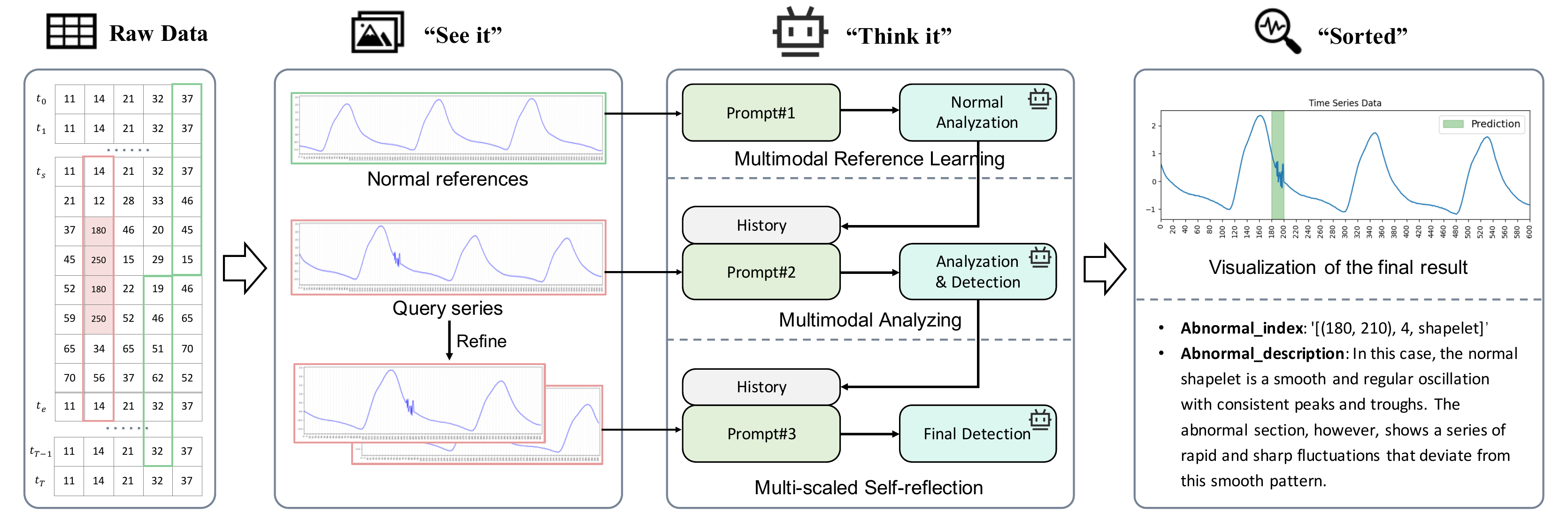}
\caption{
Our framework converts time series into images for visual interpretation (“See it”). Then, LMMs are employed to analyze the visualized time series through Multimodal Reference Learning, Multimodal Analyzing, and Multi-scaled Self-reflection, ensuring self-consistency and stability in the analysis ("Think it"). Finally, the detected anomalous intervals are processed into the output format required for TSAD, providing descriptions and possible reasons for each anomaly ("Sorted").
}
\label{fig:model}
\vspace{-10pt}
\end{figure*}

\section{Methodology}
Previous TSAD methods often rely on manual feature extraction or require large amounts of high-quality training data, along with extensive parameter tuning tailored to specific tasks. This reliance can lead to suboptimal performance and inefficiency. To overcome these limitations, we leverage the perceptual and reasoning capabilities of pretrained LMMs. Our approach transforms time series data into visual representations, enabling few-shot, high-performance, and robust anomaly detection. In this section, we first introduce the necessary preliminaries, followed by a detailed presentation of the TAMA framework.
\label{sec:method}
\subsection{Preliminary}
\label{sec:preliminary}

\textbf{Time series anomaly detection and classification .} Consider a time series data $\mathbf{x}=(x_1,\dots,x_T)\in R^T$, where $x_t$ represents the sampled value at timestamp $t$. In this paper, we focus on univariate time series, while multivariate data will be converted into multiple univariate series.
The goal of time series anomaly detection is to identify anomalous points or intervals within the time series $\mathbf{x}$. Specifically, an anomaly detection model outputs a sequence of anomaly scores $\mathbf{s} = (s_1, \dots, s_T)$, where $s_t$ indicates the anomaly score corresponding to the data point $x_t$. A higher anomaly score suggests that the model perceives a greater likelihood of the point being anomalous, while a lower score indicates a higher probability of the point being normal. By setting a threshold, the set of anomalous points can be obtained from the score sequence $\mathbf{s}$. Anomaly classification is a multi-class classification task designed to categorize identified anomalous points into specific types of anomalies. These anomaly types can provide insights into the characteristics and possibly the underlying causes of the detected anomalies.

\textbf{Preprocessing.} We follow common practice by applying mean-variance normalization to preprocess the time series, resulting in normalized data $\overline{\mathbf{x}} = (\overline{x}_1, \dots, \overline{x}_T)$, where
$
\overline{x}_t = \frac{x_t - \mu(\mathbf{x})}{\sigma(\mathbf{x})},
$
with $\mu(\cdot)$ and $\sigma(\cdot)$ denoting the mean and standard deviation functions, respectively. Then we utilize overlapped sliding windows to segment the normalized data into a collection of sequence segmentations $\mathcal{P} = \{\hat{\mathbf{x}}_1, \dots, \hat{\mathbf{x}}_l\}$, where $\hat{\mathbf{x}}_k = (\overline{x}_{k \cdot L_s}, \dots, \overline{x}_{k \cdot L_s + L_w - 1})$. $L_s$ and $L_w$ are hyperparameters representing the step size and width of the sliding window, respectively. The overlap ratio is defined as $r_o = L_s / L_w$. In this work, we choose $r_o < 1$ to allow the same segment of the sequence to be considered across multiple windows.


\textbf{Image Conversion.} To leverage the multimodal reasoning capabilities of the pretrained LMMs, we convert the time series data into graphs, serving as image modality inputs for the model. To accommodate the token limitations of the model, we impose a resolution constraint on the images. Further details can be found in Appendix \ref{sec:Appendix-suggestions}.
\subsection{Time Series Anomaly Analyzer ~(TAMA)}
The proposed TAMA framework is illustrated in Figure~\ref{fig:model}. TAMA comprises three sections that involve the participation of the LMM: \textbf{Multimodal Reference Learning}, \textbf{Multimodal Analyzing}, and \textbf{Multi-scaled Self-reflection}. Within these sections, we utilize specific prompts to guide the LMM in executing designated operations. A post-processing module is employed to integrate the model's outputs and obtain the final results. All prompts used in the experiments are detailed in Appendix \ref{sec:Appendix-Prompts}. Below, we provide a detailed explanation of TAMA's workflow.

\textbf{Multimodal Reference Learning. }
\label{sec:MRL}
This section leverages the few-shot in-context learning (ICL) capabilities of pretrained LMMs to capture the patterns of normal sequences. The model is provided with a set of normal images $\mathcal{I} = \{\mathbf{I}_i \mid i \in \{1, \dots, n_r\}\}$, where $n_r$ denotes the number of reference images. These reference images are accompanied by prompts indicating that they represent normal sequences without anomalies. The model is then tasked with generating descriptive outputs for these normal images. These descriptions will be used in subsequent interactions with the model, helping it to better focus on normal patterns and improving its ability to detect anomalous sequences.

\textbf{Multimodal Analyzing. }
\label{sec:MA}
This section utilizes the normal data patterns learned by the LMMs during reference learning to identify anomalies in new samples. Specifically, the LMMs is driven to accomplish two tasks: anomaly detection and classification.

For the $k^{\text{th}}$ sliding window, the anomaly interval detection task requires the model to output a set of anomaly intervals $\mathcal{A}_k = \{(t_s, t_e)^i\}_{i=1}^{m_k}$, where $(t_s, t_e)^i$ represents the $i$th anomaly interval, and $m_k$ is the number of detected anomaly intervals within the sliding window. We require $t_s \leq t_e$ ($t_s = t_e$ indicates a point anomaly). Note that the intervals output by the LMMs are the indices within the sliding window, which will be converted to global indices during post-processing.

Based on the results of anomaly detection, the model is then tasked with classifying each detected interval. Following \cite{lai2021revisiting}, anomalies are categorized into four types: point, shapelet, seasonal, and trend. We inform the LMMs about the characteristics of each type through natural language descriptions (see Prompt~\ref{box:multimodal_analyzing} in Appendix). The LMMs will output an anomaly set $\mathcal{Y}_k = \{y_i\}_{i=1}^{m_k}$, where $y_i$ corresponds to the anomaly classification result for the interval $(t_s, t_e)^i$.
Moreover, we also require the LMMs to provide a confidence score and an explanation for each detected anomaly interval, denoted as $\mathcal{C}_k = \{c_i\}_{i=1}^{m_k}$ and $\mathcal{T}_k = \{\text{E}_i\}_{i=1}^{m_k}$, respectively. Thus, traversing all sliding windows, the total output from the model is given by
\begin{align}
\mathcal{Z}_{\text{raw}} = \{(\mathcal{A}_k, \mathcal{Y}_k, \mathcal{C}_k, \mathcal{T}_k)\}_{k=1}^N,\nonumber
\end{align}
where $N$ is the total number of sliding windows.

\textbf{Multi-scaled Self-reflection. }
\label{MSR}
This section motivates the LMMs to correct some of its own errors, thereby enhancing the robustness and accuracy of anomaly detection. We provide the LMMs with the outputs from the previous sections, along with zoomed-in images of the detected anomaly regions. The zooming in on the anomaly areas helps prevent the model from repeatedly generating the same results due to viewing identical images. To save tokens, this operation is conducted only within the windows where some anomaly intervals are detected.


\textbf{Results Aggregation.}
In the post-processing stage, we need to consolidate the overlapped sliding windows and establish a threshold to determine the final detected anomaly points. First, we convert the indices of the anomaly intervals within the windows to a unified set of global indices. Next, we perform pointwise addition of all predicted confidence scores, i.e., if a point is detected in multiple intervals, its confidence score is the sum of the confidence scores of all these intervals. As a result, we obtain a confidence sequence $\tilde{\mathbf{c}} = (\tilde{c}_1, \dots, \tilde{c}_T)$ that enjoys the same length as the original sequence. Similarly, a pointwise anomaly classification sequence $\tilde{\mathbf{y}} = (\tilde{y}_1, \dots, \tilde{y}_T)$ is obtained, where the classification result for each point is determined by a voting mechanism from all the anomaly intervals that contain the point. Finally, by setting a threshold $c_0$, we obtain the final set of anomaly points $\mathcal{R} = \{i \ | \ 1 \leq i \leq T, \tilde c_i \geq c_0\}$.

\input{tables/full_results_new}

\section{Experiments}

In this section, we conduct extensive experiments to evaluate TAMA. The experiments include \textbf{Anomaly Detection} and \textbf{Anomaly Classification}, and \textbf{Ablation Study}. 

\textbf{Experimental Settings.} We select GPT-4o \cite{openai_gpt-4_2024} TAMA's default model, and the specific version we used is "gpt-4o-2024-05-13". To ensure the stability of TAMA and the reproducibility of the results, we set the \textbf{temperature} to 0.1 and set the \textbf{top\_p} to 0.3. Besides, we use the JSON mode of GPT-4o to facilitate subsequent result analysis. All the prompts can be found in Appendix~\ref{sec:Appendix-Prompts}. The detailed settings of image conversion has also been provided and discussed in Appendix~\ref{sec:Appendix-suggestions}.

\textbf{Datasets.} As shown in Table~\ref{tab:dataset}, we use a diverse set of real-world datasets across multiple domains for both anomaly detection and anomaly classification tasks. These domains include Web service: SMD \cite{su_robust_2019}, industry: UCR \cite{wu2021current}, NormA \cite{boniol_unsupervised_2021}, NASA-SMAP \cite{hundman_detecting_2018}, and NASA-MSL \cite{hundman_detecting_2018}, health care: ECG \cite{paparrizos_tsb-uad_2022}, and transportation: Dodgers \cite{hutchins_dodgers_2006}. All datasets are univariate except for SMD, which is originally multivariate. We convert SMD into a univariate dataset by splitting it channel-wise for our anomaly detection experiment. The full experimental results are available in Appendix~\ref{sec:Appendix-full-results-across-datasets}.

Due to the limited availability of datasets with anomaly classification labels, we created an anomaly classification dataset by combining four real-world datasets (UCR, NASA-SMAP, NASA-MSL, and NormA) with manually labeled anomaly types, along with a synthetic dataset generated using GutenTAG \cite{wenig_timeeval_2022}. To ensure the accuracy of the anomaly type annotations, cross-validation on the labels of the four real-world datasets was conducted by experts in each field. The synthetic dataset contains 7200 samples, and the anomaly types in this dataset are \textit{point}, \textit{trend}, and \textit{frequency}, as derived from the work of Lai et al. \cite{lai2021revisiting}. A visualization of each typical anomaly type is included in Appendix \ref{fig:Appendix-Anomaly_Visualizaion}. 


\subsection{Anomaly Detection}
\label{sec:AnomalyDetection}
In this section, we evaluate the anomaly detetcion capability of TAMA.


\input{tables/dataset}

\textbf{Baselines.} The baseline models used in our experiments include both machine learning algorithms ~(IF \cite{liu_isolation_2008}, LOF \cite{huang_lof-based_2013}) and deep learning ~(TranAD \cite{tuli2022tranad}, GDN \cite{deng_graph_2021}, MAD\_GAN \cite{li_mad-gan_2019}, MSCRED \cite{zhang_deep_2018}, MTAD\_GAT \cite{zhao_multivariate_2020}, OminiAnomaly \cite{su_robust_2019}, USAD \cite{audibert_usad_2020} and TimesNet \cite{wu_timesnet_2023}).
\footnote{All deep learning baseline models are from the repository of TranAD~(VLDB'22, Github: https://github.com/imperial-qore/TranAD)}
Besides, the SIGLLM~\cite{alnegheimish_large_2024} is a baseline model based on the LLM. We reproduce it with GPT-4o.
All baseline models has been run with the default configurations. For those datasets without default configurations, we managed to optimize the performance by searching the best parameters.

\textbf{Metrics.} Following the mainstream of TSAD, we evaluated TAMA and other baselines by the point-adjusted \emph{F1} \cite{xu_unsupervised_2018}. Point adjustment (PA) is a widely used adjustment method in TSAD tasks \cite{kim_towards_2021,blazquez-garcia_review_2021}, but it can significantly overestimate the models' performances (especially F1), making it a subject of much debate. A detailed explanation and further discussion of  are available in Appendix~\ref{sec:Appendix-PA-metrics}.

To fully assess how the models' performance at different levels of sensitivity and specificity, we also adopted two threshold-agnostic metrics, namely \emph{AUC-PR} and \emph{AUC-ROC}. We would like to highlight \emph{AUC-PR} as it is more robust to scenarios with highly imbalanced classes (like TSAD) comparing to \emph{AUC-ROC} \cite{sofaer_area_2019}. 

\begin{figure*}[t]
    \centering
    \includegraphics[width=\linewidth]{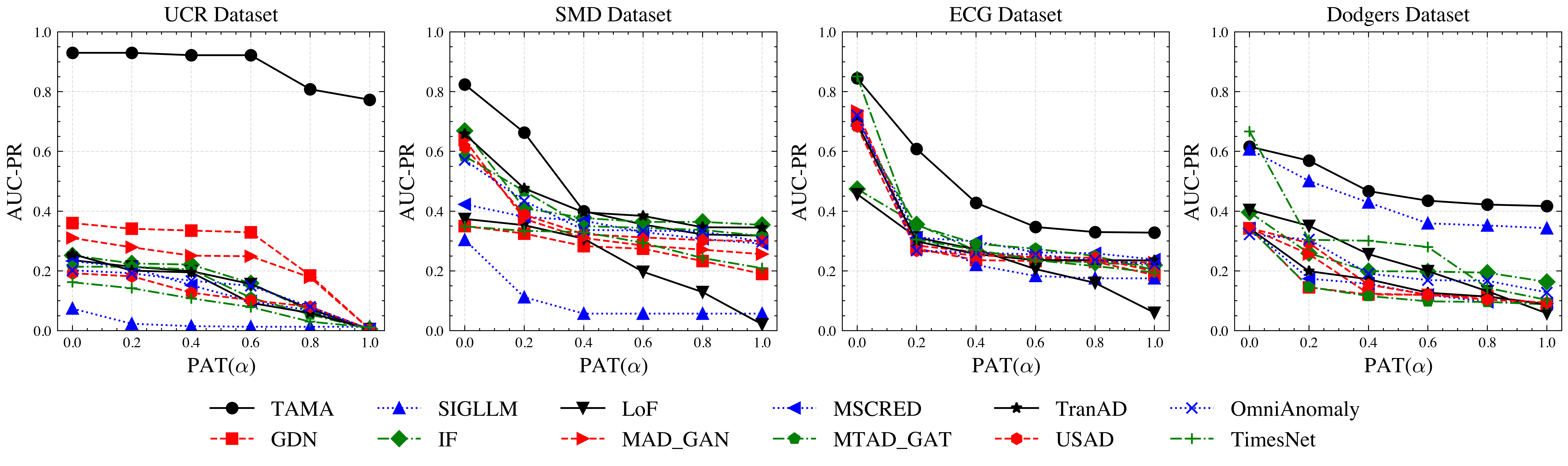}
    \vspace{-20pt}
    \caption{The \emph{AUC-PR} of all models at various point-adjustment threshold $\alpha$~(PAT, defined in Appendix~\ref{sec:Appendix-PA-metrics}). 
    } 
    \label{fig:PAT}
\end{figure*}

\textbf{Main Results.} The experiment results are shown in Table~\ref{tab:main_results}. Each metric in the table contains two value: \emph{mean} and \emph{maxima}. The \emph{mean} refers to the average of all sub-series, while the \emph{maxima} represents the best result among all sub-series.
For the \emph{maxima} value, our method~(TAMA) generates comparable results as other baseline models, even better in some datasets. TAMA outperforms almost all baseline models across every dataset in terms of \emph{mean} metrics, especially on industry and transportation datasets. The exceptional performance of the \emph{mean} suggests that our approach exhibits superior stability across a broader range of data. 
In contrast, most baseline models are sensitive to their hyper-parameters, leading to limited generalization.

\textbf{Self-reflection.} 
In order to improve the stability, we add the self-reflection to TAMA. The results are presented in Table~\ref{tab:main_results}. \textbf{TAMA*} represents the method without self-reflection. We can find that in most datasets, the \emph{maxima} of \textbf{TAMA*} is very close to that of \textbf{TAMA}. However, there are obvious drops of \emph{mean} in most datasets, especially in NASA-SMAP. These drops demonstrate the efficacy of self-reflection in stability.

\textbf{The impact of point-adjustment.} 
Under the point-adjustment, if a single point is predicted correctly in an interval, the entire interval is considered to be right, which greatly overestimates the model's performance. 
To address this, we re-evaluate all models using point-adjustment with a threshold $\alpha$~(PAT, defined in Appendix~\ref{sec:Appendix-PA-metrics}).

The results are shown in Figure~\ref{fig:PAT} (full results in Appendix~\ref{sec:Appendix-full-results-PAT}.). We only present results of four datasets due to the limitation of the space, the full results are available in Appendix~\ref{sec:Appendix-full-results-PAT}. 
The UCR, SMD, ECG, and Dodgers datasets originate from the domains of industry, web services, healthcare, and transportation, respectively.

Based on the experimental results, we can find that the performance of all models declines as the PAT increases, indicating that full point-adjustment~($\alpha$=0) has overestimated the model performance. However, compared to other baseline methods, TAMA demonstrates excellent performance across all PATs, showing that TAMA can identify the anomaly more accurately.

Overall, the results of anomaly detection in Table~\ref{tab:main_results} reveals that the LMM is an excellent and stable anomaly detector. The ability of multimodal reasoning can be utilized in analyzing the image of time series data and achieve an impressive standard.

\subsection{Anomaly Classification}
\label{sec:AnomalyClassification}
In practical applications, it is preferable not only to detect anomaly intervals but also to provide a brief classification indicating their causes. To fully demonstrate the strong reasoning capabilities of the proposed framework and enhance the interpretability of detection results, we conduct classification on the anomaly data.


\input{tables/classification}

The overall results presented in Table~\ref{tab:o_class_r} indicate that TAMA, guided by the provided prompts (outlined in Appendix \ref{sec:Appendix-Prompts}), demonstrates a reliable understanding of each type of anomaly and can accurately classify most anomalies, with the exception of seasonal anomalies. TAMA performs exceptionally well in classifying shapelet anomalies, suggesting that it effectively captures the shape of the input sequences. However, it is evident that the framework struggles with seasonal anomalies. 
We interpret this difficulty as stemming from a lack of relevant materials in the LMM's pre-training stage, which results in a weak understanding of concepts such as "seasonality" or "frequency". More discussion on TAMA's behavior to each anomaly type is included in Appendix \ref{sec:Appendix-Type-Detection}.  

\input{tables/ablation-LMM}

\subsection{Ablation Study}
There are many hyperparameters affecting the performance of our framework. In this section, some ablation experiments are conducted to evaluate the impact of each hyperparameter, including \textbf{LMM Selection}, \textbf{Reference Number} and \textbf{Window Size}. 

\textbf{LMM Selection.}
In this section, we valid that the design of TAMA enhances the capability of LMMs in anomaly detection task. We conduct experiments on the UCR dataset with various LMMs, including GPT-4o, GPT-4o-mini, Gemini-1.5-pro, Gemini-1.5-flash, and Qwen-vl-max. For each LMM, we conduct experiments both with ~(\textit{+\textbf{TAMA}}) and without ~(\textit{Naive}) TAMA framework. Due to budget constraints, we only conduct this experiment on the UCR dataset. However, we believe the experimental results can, to some extent, reflect real-world scenarios.

The experimental results are presented in the Table~\ref{tab:ablation-LLMs}. To better evaluate the efficacy of our framework~(TAMA), we use the original \emph{AUC-PR} without point-adjustment as the metric in this experiment. The findings reveal that all LMMs exhibit a substantial enhancement in performance on the UCR dataset following their integration into TAMA. This not only validates that TAMA improves the LMMs' abilities in anomaly detection but also confirms the generalizability of TAMA's framework.

\textbf{Reference Number.} As mentioned earlier in Section~\ref{sec:MRL}, we provide some normal images $\mathbf{I}=\{I_i|i\in[1,N]\}$ as references to help the LMM learn the distribution of normal data. In this ablation experiment, we investigate the impact of the number of reference images $N$. The ablation experiment is conducted on UCR and NASA-SMAP datasets. The ablation experiment is conducted on the UCR and NASA-SMAP datasets. To more clearly highlight the differences in performance, we used metrics without PA. Moreover, to avoid interference from other modules and more realistically demonstrate the role of Multimodal Reference Learning, self-reflection is not added in this ablation experiment.

\begin{figure}[t]
    \centering
    \includegraphics[width=\columnwidth]{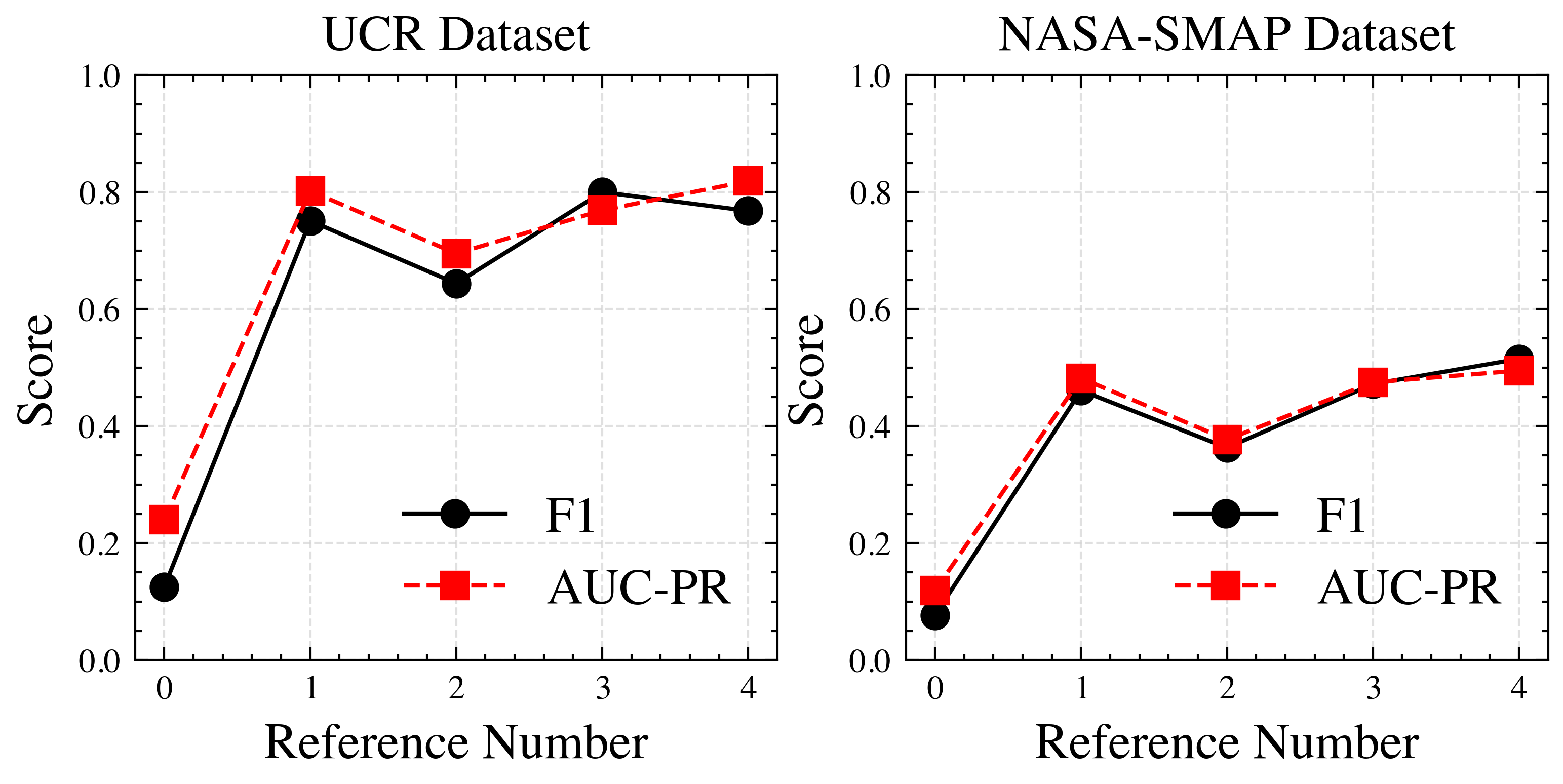}
    \vspace{-20pt}
    \caption{
    Results (without PA) of reference number ablation experiments.
    }
    \label{fig:ab-normal-reference-results}
    \vspace{-14pt}
\end{figure}

As shown in the Figure \ref{fig:ab-normal-reference-results}, when the Reference Number is set to 0, which indicates that no reference is provided in the framework, the performance has a significant drop. As the number of Reference Number increases, although there are some fluctuations, it remains relatively stable, and the performance is considerably improved to the case without references.

Our results show that using Reference Learning is highly necessary, as it can significantly improve the method's performance and stability. At the same time, from the experimental results, we also find that the quantity of references, from 1 to 4, does not seem to have a clear relationship with the final results. However, in case of the random sampling of normal data, setting the Reference Number to 3 can enhance overall robustness. 

\begin{figure}[t]
    \centering
    \includegraphics[width=\columnwidth]{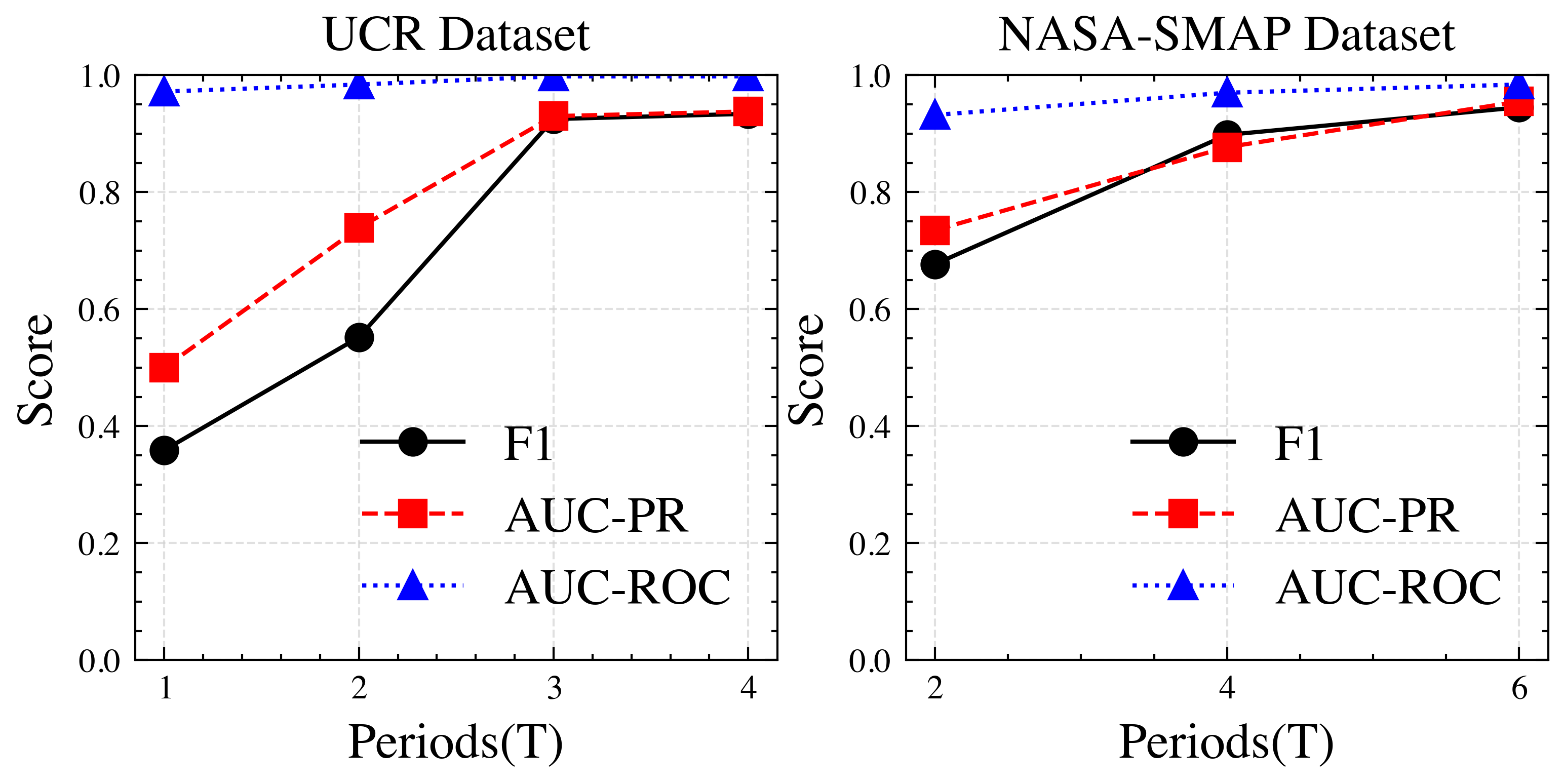}
    \vspace{-20pt}
    \caption{Results of window size ablation experiments. For the period of two datasets, $T_{UCR} \approx 200$, $T_{NASA-SMAP} \approx 100$}
    \label{fig:ab-ws-results}
    \vspace{-15pt}
\end{figure}

\textbf{Window Size.} In TAMA, as it shown before (in Section \ref{sec:preliminary}), we use sliding window in the data pre-processing stage. To accommodate different time series data with varying periods, we report the window size in multiples of the data period.


The results presented in Figure~\ref{fig:ab-ws-results} reveal that the performance of our method is positively correlated with the window size. This is because the LMMs struggles to identify periodic patterns when given only single-period images, resulting in incorrectly classifying periodic features and truncated features as anomalies. Therefore, we ultimately set the window size to approximately $3T$ for the experiments detailed in Section \ref{sec:AnomalyDetection}.

\section{Discussion}
In this section. we reflect on the design of TAMA and seek to answer the following research questions:

\textbf{RQ1}: Is the image modality better than text modality for LMMs in the time series anomaly detection task?~(Section~\ref{sec:discussion-modality})

\textbf{RQ2}: How does the additional information on images affect the LMM's analysis?~(Section~\ref{sec:discussion-image})

\textbf{RQ3}: Dose the LMM truly learn the reference data during the multimodal Rederence Learning.~(Section~\ref{sec:discussion-MRL})

\subsection{Modality}
\label{sec:discussion-modality}
To investigate whether the success of our framework comes from the more advanced model~(GPT-4o) we choose or the image-modality. we conduct an  experiment in NASA-MSL and NASA-SMAP datasets, which both are real-world datasets and have more complex patterns than the UCR dataset.

The results are shown in Table~\ref{tab:ablation-modality}. We choose different modalities for testing while trying to keep the prompts and procedures as the same. To maintain fairness, we do not add the self-reflection in TAMA. Meanwhile, we also include the results of SIGLLM~\cite{alnegheimish_large_2024}, which uses \emph{Mistral-7B}, as a reference. Compared to methods using text-modality, TAMA~~(Image), which use image-modality, has made significant improvement, with a 37.9\% increase on NASA-MSL and a 36.9\% increase on NASA-SMAP. This result indicates that for anomaly detection tasks, allowing the model to "see" time series data~~(using image modality) is more beneficial. It also demonstrates that multimodal reasoning has tremendous promise in time series anomaly detection tasks.

\input{tables/ablation-modality}

\subsection{Additional Information in Images}
\label{sec:discussion-image}
The transformation of raw data into visual formats, such as images, adds crucial information, including plot orientation and auxiliary lines. This study investigates how these elements influence TAMA's performance in identifying abnormal intervals based on plot scales. We conducted two experiments: the first involved rotating images by 90 degrees before inputting them into TAMA, while the second examined the impact of auxiliary lines, which are perpendicular to the x-axis and align with the scale to aid in locating data points.

Both experiments are performed on the UCR and NASA-SMAP datasets. Results are presented in Table \ref{tab:ablation-rotation}, where TAMA represents the original model, TAMA-R indicates performance with rotated images, and TAMA-A reflects performance without auxiliary lines. We evaluated using the AUC-PR without point adjustment. The findings demonstrate a notable decline in TAMA-R's performance with rotated images, suggesting that the LMMs are sensitive to image orientation. Despite the rotation of axis is disturbed in prompts, the LMM struggles to interpret rotated images accurately, leading to reduced anomaly detection. In contrast, TAMA-A experiences only a slight performance decrease across both datasets, indicating that LMMs can better identify abnormal intervals when auxiliary lines are present.

These experiments reveal that LMMs perceive time series images similarly to humans--- uxiliary lines enhance anomaly localization accuracy, while image rotation negatively affects performance. This sensitivity may result from the tokenizer's responsiveness to orientation or insufficient training data and guidance.
\input{tables/ablation-rotation}

\input{tables/ablation-NormalReference}

\subsection{Multimodal Reference Learning}
\label{sec:discussion-MRL}
We provide a set of images for LMMs, helping LMMs to better focus on normal pattern and improving the ability to detect anomalous intervals (Section \ref{sec:MRL}). To study whether the LMM truly learns from the reference images, we conduct this experiment on the UCR and NASA-SMAP datasets, replacing the normal data with abnormal data, and Multi-scaled Self-reflection is not enabled in this experiment. The results are presented in Table~\ref{tab:ablation-NR}. In the table, \textit{normal} refers to using the normal data as the reference data, while \textit{abnormal} indicates using abnormal data as reference data. The experimental results under different reference data conditions are  evaluated by the \emph{AUC-PR} without point-adjustment. We can find that \textit{normal} performs better than \textit{abnormal} on both UCR and NASA-SMAP datasets, showing that the content for reference learning can notably impact the model's performance, which suggests the LMM can truly learn normal representation from the reference data.


\section{Conclusion}
In this paper, we introduced TAMA, a novel framework that leverages large multimodal models for effective time series anomaly analysis. Comprehensive evaluation across multiple metrics demonstrates that TAMA not only surpasses state-of-the-art methods but also provides valuable semantic classifications and insights into detected anomalies. By converting time series data into visual representations, we have, for the first time, applied large multimodal models to this domain, enabling stronger generalization and more robust interpretative analysis. In summary, TAMA represents a significant advancement in anomaly detection methodologies, with practical implications for real-world applications and new opportunities for future research in multidimensional anomaly detection.

However, some limitations should be noted. Firstly, the proposed approach primarily relies on the pre-trained LMMs without fine-tuning. Additionally, in this work, we only consider univariate time-series anomaly detection tasks, whereas in real-world scenarios, it’s necessary to incorporate multiple time series for comprehensive judgment. In future works, we plan to explore deploying large models locally and fine-tuning them to achieve better performance and enhanced data security. Furthermore, we are considering the incorporation of multidimensional time series anomaly detection in our subsequent research efforts.





\bibliographystyle{ACM-Reference-Format}
\bibliography{refs, references}

\clearpage
\appendix
\section{Appendices}
\label{sec:appendix}

\subsection{Prompts}
\label{sec:Appendix-Prompts}

The design of prompts is based on 
the documentation of OpenAI\footnote{https://platform.openai.com/docs/guides/prompt-engineering/strategy-write-clear-instructions}. Writing the steps out explicitly can make it easier for the model to follow them. In our task, we separate the whole task into three specific tasks: \textbf{Multimodal Reference Learning}~(see Prompt~\ref{box:multimodal_reference}), \textbf{multimodal Analyzing}~(see Prompt~\ref{box:multimodal_analyzing}) and \textbf{Multi-scaled Self-reflection}~(see Prompt~\ref{box:multisclaed_self_reflection}). Besides, we also provide some background information, such as siliding windows and additional information of images. With the JSON mode output of GPT-4o, it is very convenient for us to process the output results, requiring a detailed description of the output format in prompts. Based on our practical experience, we find that clear descriptions and a structured format significantly are very helpful for LMM to understand. 

\subsection{Point Adjustment Metrics}
\label{sec:Appendix-PA-metrics}

Given a set of real anomalous intervals $\mathcal{A}_{\text{T}} = \{(t_s, t_e)^i_{\text{T}}\}_{i=1}^{m_T}$ and a set of predicted anomalous intervals $\mathcal{A}_{\text{P}} = \{(t_s, t_e)^i_{\text{P}}\}_{i=1}^{m_P}$, the point-adjusted prediction $\mathcal{A}_{\text{PA}}$ is defined as:
\begin{equation}
    \mathcal{A}_{\text{PA}} = \mathcal{A}_{\text{P}}\cup\{t|t\in(t_s, t_e)^i_{\text{T}},
    |(t_s, t_e)^i_{\text{T}} \cap (t_s, t_e)^j_{\text{P}}|>0\},
    \label{eq:A-PA}
\end{equation}
where the $\mathcal{A}_{\text{PA}}$ is a set of points, $m_T$ and $m_P$ refer to the total number of real anomalous intervals and the total number of the LLM's prediction, respectively. 
After point-adjustment, the point-adjusted \emph{Recall}, \emph{Precision} and \emph{F1} can be calculated as:
\begin{align}
    R &= \emph{Recall}~(\mathcal{A}_{\text{T}},\mathcal{A}_{\text{PA}}), \label{eq:pa-recall}\\
    P &= \emph{Precision}~(\mathcal{A}_{\text{T}},\mathcal{A}_{\text{PA}}), \label{eq:pa-precision} \\
    \emph{F1} &= {2~(R+P)} / {~(R \cdot P)}, \label{eq:pa-f1}   
\end{align}
The point-adjustment with threshold $\alpha$ is defined as:
\begin{equation}
\begin{aligned}
    \mathcal{A}_{\text{PA}}(\alpha) &= \mathcal{A}_{\text{P}} \cup \{t | t \in (t_s, t_e)^i_{\text{T}}, \\
    & \quad |(t_s, t_e)^i_{\text{T}} \cap (t_s, t_e)^j_{\text{P}}| > \alpha \cdot L((t_s, t_e)^i_{\text{T}})\}.
\end{aligned}
\end{equation}

where $\alpha$ refers to the point-adjustment threshold~(PAT) from 0 to 1, where 0 represents full point-adjustment $\mathcal{A}_{\text{PA}}$ and 1 represents original prediction $\mathcal{A}_{\text{P}}$, and the $L((t_s, t_e)^i_{\text{T}})$ refers to the length of $(t_s, t_e)^i_{\text{T}}$. 

\subsection{Some Suggestions about TAMA}
\label{sec:Appendix-suggestions}
In this paper, we propose a framework named TAMA to utilize the LMM to analyze time series images. However, we have tried multiple versions and gained valuable practical experience during the development process.
Based on our practical experience, we provide some suggestions.
\begin{itemize}
    \item To better parse the output results, choosing the LMM which supports JSON mode output or structured output can be very convenient. If the LMM does not support these output format, we can use GPT-4o, which supports structured output, to format the output text.
    \item Assume the period of series data is $T$, it is recommended to set the sliding window length to at least $3T$.
    \item The LMM marks the interval with anomaly based on the scale of the plot. Therefore, the scale of axis should be clear enough. However, the rotation of scale does not matter.
    \item Grid-like auxiliary lines can be added to enhance the accuracy of the anomaly intervals output by the LMM.
    \item According to the documentation of OpenAI, in order to use high revolution mode, the figure size should not larger than 2000x768 pixels. All images in TAMA will be limited to this size.
\end{itemize}

\begin{NewBox}*[hb]{box:multimodal_reference}{Multimodal Reference Learning Prompt}

<Background>: 
I have a long time series data with some abnormalities. I have converted the data into plots and I need your help to find the abnormality in the time series data.
This task contains two parts:

\begin{itemize}
    \item[-] "Task1": I will give you some "normal reference" time series data slices without any abnormality. And you need to extract some valuable information from them to help me find the abnormality in the following time series data slices.
    \item[-] "Task2": I will give you some time series data slices with some abnormalities. You need to find the abnormality in them and provide some structured information.
\end{itemize}
besides, I will offer you some background information about the data plots:
\begin{itemize}
    \item[-] The horizontal axis represents the time series index.
    \item[-] The vertical axis represents the value of the time series.
    \item[-] all normal reference data slices are from the same data channel but in different strides. Therefore, some patterns based on the position, for example, the position of peaks and the end of the plot, may cause some confusion.
    \item[-] all normal references are slices of the time series data with a fixed length and the same data channel. Therefore the beginning and the end of the plot may be different but the pattern should be similar.
\end{itemize}

<Task>: 
Now we are in the "Task1" part: I will give you some "normal reference" time series data slices without any abnormality. And you need to extract some valuable information from them to help me find the abnormality in the following time series data slices.

<Target>: 
Please help me extract some valuable information from them to help me find the abnormality in the following time series data slices.
The output should include some structured information, please output in JSON format:
\begin{itemize}
    \item[-] normal\_pattern ~(a 300-400 words paragraph): Try to describe the pattern of all "normal references" . All normal reference data slices are from the same data channel but in different strides. The abnormal pattern caused by truncation might be found at the beginning and end of the sequence, do not pay too much attention to them. The description should cover at least the following aspects: period, stability, trend, peak, trough, and other important features.
\end{itemize}
\end{NewBox}

\begin{NewBox}*[ht]{box:multisclaed_self_reflection}{Multi-scaled Self-reflection}
<Background>: 
I have a long time series data with some abnormalities. I have converted the data into plots and I need your help to find the abnormality in the time series data.
There has been a response from another assistant, but I am not sure about the prediction. I need your help to double check the prediction.
Besides, I will offer you some background information about the data plots: 
\begin{itemize}
    \item [-] The horizontal axis represents the time series index.
    \item [-] The vertical axis represents the value of the time series.
    \item [-] all normal reference data slices are from the same data channel but in different strides. Therefore, some patterns based on the position, for example, the position of peaks and the end of the plot, may cause some confusion.
    \item [-] all normal references are slices of the time series data with a fixed length and the same data channel. Therefore, the beginning and the end of the plot may be different, but the pattern should be similar.
\end{itemize}
<Task>:
Now, I will give you some "normal reference" and you are expected to double check the prediction of the abnormality in the given data.

<Target>:
The prediction of another assistant contains some information as follows: 
    \begin{itemize}
        \item [-] abnormal\_index: The abnormality index of the time series. The output format should be like "[~(start1, end1)/confidence\_1/abnormal\_type\_1, ~(start2, end2)/confidence\_2/abnormal\_type\_2, ...]", if there are some single outliers, the output should be "[~(index1)/confidence\_1/abnormal\_type\_1, ~(index2)/confidence\_2/abnormal\_type\_2, ...]",if there is no abnormality, you can say "[]".
        \item [-] abnormal\_description: Make a brief description of the abnormality, why do you think it is abnormal?
    \end{itemize}
        Based on the "nomral reference" I gave you, please read the prediction above and double check the prediction. If you find any mistakes, please correct them. The output should include some structured information, please output in JSON format:
    \begin{itemize}
        \item [-] corrected\_abnormal\_index ~(string, the output format should be like "[~(start1, end1)/confidence\_1/abnormal\_type\_1, ~(start2, end2)/confidence\_2/abnormal\_type\_2, ...]", if there are some single outliers, the output should be "[~(index1)/confidence\_1/abnormal\_type\_1, ~(index2)/confidence\_2/abnormal\_type\_2, ...]",if there is no abnormality, you can say "[]". The final output should be mixed with these three formats.): The abnormality index of the time series. There are some requirements:
            \begin{itemize}
            \item [+] 1. you should check each prediction of the abnormal\_type and make sure it is correct based on the abnormality index. If there is a incorrect prediction, you should remove it.
            \item [+] 2. you should check each prediction of the abnormal\_index according to the image I gave to you. If there is an abnormality in image but not in the prediction, you should add it. The format should keep the same as the original prediction.
            \end{itemize}
        \item [-] The reason why you think the prediction is correct or incorrect. ~(a 200-300 words paragraph): Make a brief description of your double check, why do you think the prediction is correct or incorrect?
    \end{itemize}
\end{NewBox}

\begin{NewBox}*[t]{box:multimodal_analyzing}{Multimodal Analyzing Prompt}
<Background>: 
I have a long time series data with some abnormalities. I have converted the data into plots and I need your help to find the abnormality in the time series data.
This task contains two parts: 
\begin{itemize}
    \item [-] "Task1": I will give you some "normal reference" time series data slices without any abnormality. And you need to extrace some valuable information from them to help me find the abnormality in the following time series data slices.
    \item [-] "Task2": I will give you some time series data slices with some abnormalities. You need to find the abnormality in them and provide some structured information.
\end{itemize}
Besides, I will offer you some background information about the data plots: 
\begin{itemize}
    \item [-] The horizontal axis represents the time series index.
    \item [-] The vertical axis represents the value of the time series.
    \item [-] all normal reference data slices are from the same data channel but in different strides. Therefore, some patterns based on the position, for example, the position of peaks and the end of the plot, may cause some confusion.
    \item [-] all normal references are slices of the time series data with a fixed length and the same data channel. Therefore the beginning and the end of the plot may be different but the pattern should be similar.
\end{itemize}

<Task>: 
In "Task1" part, you have already extracted some valuable information from the "normal reference" time series data slices. You can use them to help you find the abnormality in the following time series data slices.
Now we are in "Task2", you are expected to detect the abnormality in the given data. 

<Target>: 
Please help me find the abnormality in this time series data slice and provide some structured information.
The output should include some structured information, please output in JSON format: 
    \begin{itemize}
        \item [-] abnormal\_index ~(the output format should be like "[~(start1, end1)/confidence\_1/abnormal\_type\_1, ~(start2, end2)/confidence\_2/abnormal\_type\_2, ...]", if there is no abnormality, you can say "[]". The final output should be mixed with these three formats.): The abnormality index of the time series. There are some requirements:
        \begin{itemize}
            \item[+]  There may be multiple abnormalities in one stride. Please try to find all of them. Pay attention to the range of each abnormality, the range should cover each whole abnormality in a suitable range. 
            \item[+] Since the x-axis in the image only provides a limited number of tick marks, in order to improve the accuracy of your prediction, please try to estimate the coordinates of any anomaly locations based on the tick marks shown in the image as best as possible. 
            \item[+] all normal reference data slices are from the same data channel but in different strides. Therefore, some patterns based on the position, for example, the position of peaks and the end of the plot, may cause some confusion. 
            \item[+] abnormal\_type~(answer from "global", "contextual", "frequency", "trend", "shapelet"): The abnormality type of the time series, choose from [global, contextual, frequency, trend, shapelet]. The detailed explanation is as follows: 
                \begin{itemize}
                \item [+] global: Global outliers refer to the points that significantly deviate from the rest of the points. Try to position the outliers at the center of the interval. 
                \item [+] contextual: Contextual outliers are the points that deviate from its corresponding context, which is defined as the neighboring time points within certain ranges. Try to position the outliers at the center of the interval.
                \item [+] frequency: Frequency outliers refer to changes in frequency, the basic shape of series remains the same. Please focuse on the horizontal axis to find the frequency anomalies.
                \item [+] trend: Trend outliers indicate the subsequences that significantly alter the trend of the time series, leading to a permanent shift on the mean of the data. Mark the intervals where the mean of the data significantly changes.
                \item [+] shapelet: Shapelet outliers refer to the subsequences with totally different shapes compared to the rest of the time series.
                \end{itemize}
        \end{itemize}
    \item [-] confidence ~(integer, from 1 to 4): The confidence of your prediction. The value should be an integer between 1 and 4, which represents the confidence level of your prediction. Each level of confidence is explained as follows:
        \begin{itemize}
        \item [+] 1: No confidence: I am not sure about my prediction
        \item [+] 2: Low confidence: Weak evidence supports my prediction 
        \item [+] 3: medium confidence: strong evidence supports my prediction
        \item [+] 4: high confidence: more than 95
        \item [+] based on the provided abnormal\_type, you should double check the abnormal\_index.
        \end{itemize}
    \item [-] abnormal\_description ~(a 200-300 words paragraph): Make a brief description of the abnormality, why do you think it is abnormal? 
    \item [-] abnormal\_type\_description ~(a 200-300 words paragraph): Make a brief description of the abnormality type for each prediction, why do you think this type is suitable for the abnormality?
    \end{itemize}
Last, please double check before you submit your answer.
\end{NewBox}


\subsection{Full Results of Anomaly Detection across All Datasets}
\label{sec:Appendix-full-results-across-datasets}
In this section, we present the full results of all datasets in Table~\ref{tab:appendix_main_results}. Due to the limitation of the space, we only present some of them in the main body. Meanwhile, we also present the variance in this table. Most of datasets contain more than one sub-sequence, to fully present and compare the performance, we evaluate all metrics in all sub-sequence and calculate three values: \emph{mean}, \emph{variance} and \emph{maxima}. In this table, \emph{mean} and \emph{variance} are formated as "\emph{mean} $\pm$ \emph{variance}".

\input{tables/full_results}
\subsection{Full Results of the PAT Experiment}
\label{sec:Appendix-full-results-PAT}
\begin{figure*}[t]
    \centering
    \includegraphics[width=\linewidth]{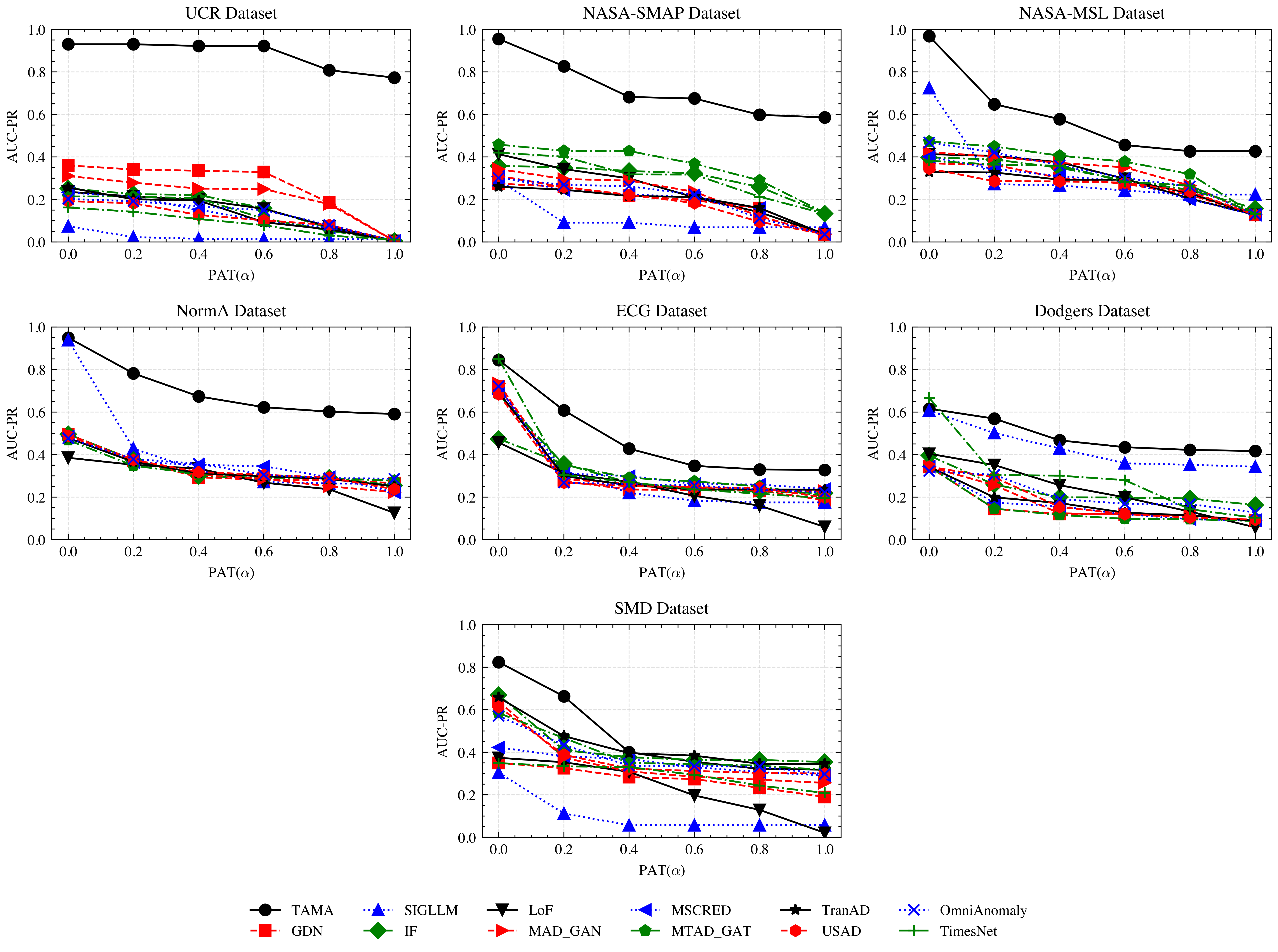}
    \caption{
    The full \emph{AUC-PR} results of all models across all datasets at various point-adjustment threshold $\alpha$~(PAT, see~Appendix~\ref{sec:Appendix-PA-metrics}).
    } 
    \label{fig:Appendix-Full-PAT}
\end{figure*}

In Section~\ref{sec:AnomalyDetection}, in order to study the impact of point-adjustment, we re-evaluate the results using the point-adjustment with a threshold $\alpha$~(See~Appendix~\ref{sec:Appendix-PA-metrics}). Due to the limitation of the space, we only present results of some datasets in the main body. The full results are presented in Figure~\ref{fig:Appendix-Full-PAT}. As the figure presented, our framework achieves outstanding \emph{AUC-PR} across all datasets at various $\alpha$, showing that our framwork has better robustness and stability.

\subsection{Performance Evaluation of Type-Specific Anomaly Detection Methods}
\label{sec:Appendix-Type-Detection}
\input{tables/type-wise-eval}
Table \ref{tab:appendix_wise_eval} presents the type-specific anomaly detection performances. To maintain readability, only the F1-score without point adjustment is reported. The results highlight TAMA's outstanding performance in identifying pattern anomalies, including shapelet, seasonal, and trend types, while most baseline models struggle in this aspect without point adjustment. For instance, on the UCR-shapelet dataset, TAMA outperformed the second-best detector (GDN) by a substantial margin of 293\% in terms of the mean F1-score. This superiority stems from TAMA's inherent ability to detect anomalous intervals. However, this characteristic may lead to lower F1-scores in the detection of point anomalies. In the synthetic dataset we generated, labels for point anomalies were strictly defined. While TAMA's interval detection always encompassed the ground-truth anomalies, it also produced a significant number of false positives.

\subsection{Visualization of anomaly classification}
\begin{figure*}[t]
    \centering
    \includegraphics[width=\linewidth]{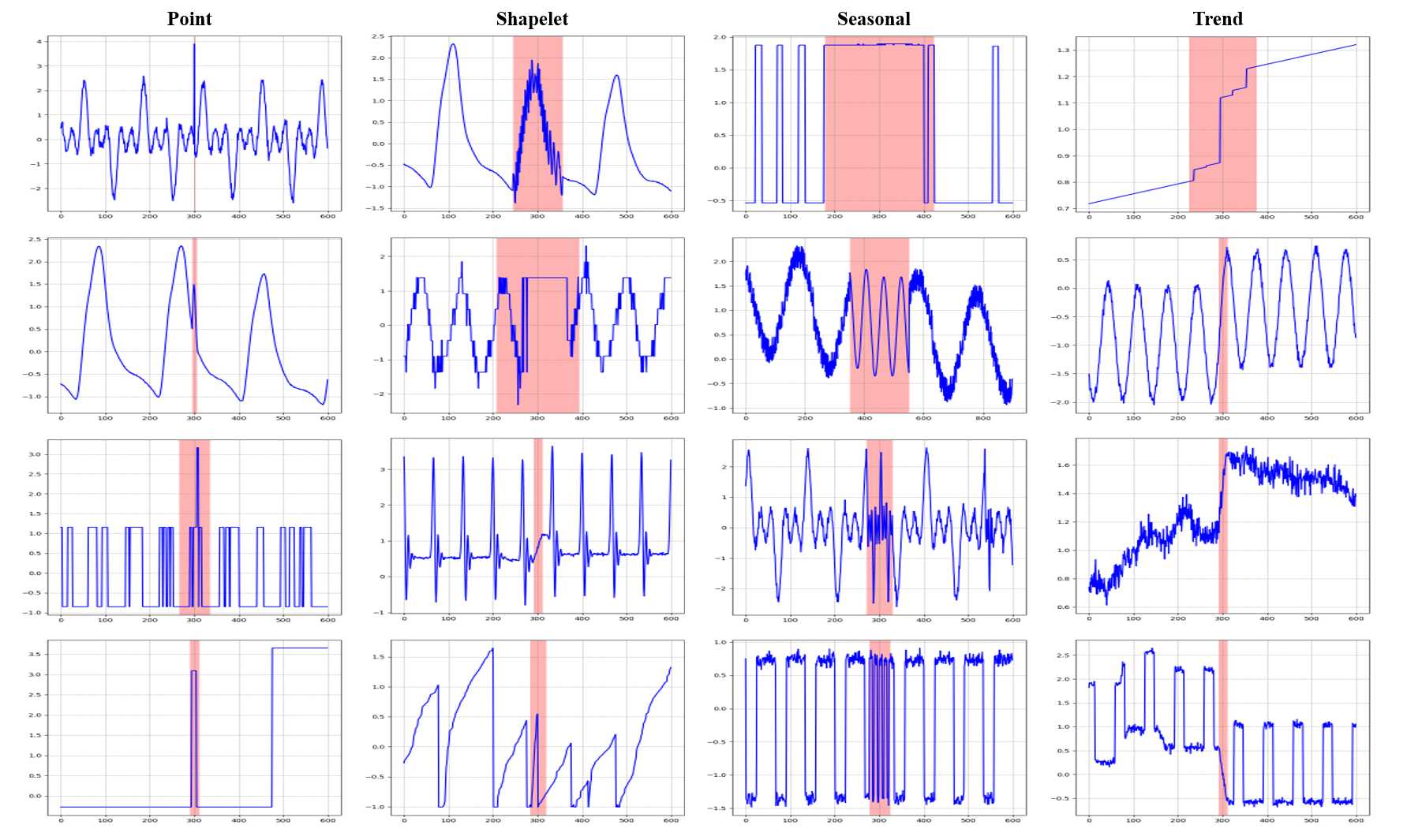}
    \caption{Visualization of anomalies. Each row displays sequences from different datasets that contain the same type of anomaly. } 
    \label{fig:Appendix-Anomaly_Visualizaion}
\end{figure*}

In Section~\ref{sec:AnomalyClassification}, we make a new dataset for anomaly classification by labeling some real-world datasets and generating sequence. We also provide some visualization of these anomalies to better understand the different types of anomalies. The visualization of anomaly classification is shown in Figure~\ref{fig:Appendix-Anomaly_Visualizaion}. The dataset contain four classification: Point, Shapelet, Seasonal and Trend, which are referenced from the work~\cite{lai2021revisiting}.

\label{sec:Appendix-VS-AC}

\end{document}

%% file: tables/full_results_new.tex
\begin{table*}[t!]
    \caption{Quantitative results across six datasets use metrics point-adjusted \emph{F1\%}, \emph{AUC-PR\%}, and \emph{AUC-ROC\%}. Best and second-best results are in bold and underlined, respectively. \textbf{TAMA} represents our framework, and \textbf{TAMA*} represents our framework without self-reflection. Each unit in the table contains two value: \emph{mean} and \emph{maxima} of all series.
    } %
    \label{tab:main_results}
    \vspace{-5pt}
    \centering
    \resizebox{0.9\textwidth}{!}{
    \renewcommand{\multirowsetup}{\centering}
    \begin{tabular}{c|C{7pt}C{20pt}C{7pt}C{20pt}C{7pt}C{20pt}|C{7pt}C{20pt}C{7pt}C{20pt}C{7pt}C{20pt}|C{7pt}C{20pt}C{7pt}C{20pt}C{7pt}C{20pt}}
    \toprule
    Dataset & \multicolumn{6}{c|}{\textbf{UCR}} & \multicolumn{6}{c|}{\textbf{NASA-SMAP}}& \multicolumn{6}{c}{\textbf{NASA-MSL}}\\
    \midrule
    Metric& \multicolumn{2}{c|}{F1\%}& \multicolumn{2}{c|}{AUC-PR\%}& \multicolumn{2}{c|}{AUC-ROC\%}& \multicolumn{2}{c|}{F1\%}& \multicolumn{2}{c|}{AUC-PR\%}& \multicolumn{2}{c|}{AUC-ROC\%}& \multicolumn{2}{c|}{F1\%}& \multicolumn{2}{c|}{AUC-PR\%}& \multicolumn{2}{c}{AUC-ROC\%}\\
    \midrule
    IF & 24.7 & 77.3 & 37.7 & 44.6 & 24.4 & 25.3 & 54.2 & 94.2 & 58.9 & 77.1 & 65.0 & 87.7 & 47.6 & 88.6 & 53.6 & \underline{80.4} & 68.7 & 88.7 \\
    LOF & 42.8 & \textbf{100} & 35.6 & 50.0 & 92.8 & \textbf{99.9} & 62.2 & \textbf{100} & 43.4 & 61.4 & 60.1 & \underline{99.9} & 36.4 & 66.8 & 44.5 & 66.0 & 58.6 & 99.8 \\
    TranAD & 38.2 & 93.7 & 30.9 & 51.0 & 77.0 & \textbf{99.9} & 59.0 & 99.6 & 36.8 & 73.9 & 74.4 & \textbf{100} & 64.6 & 99.1 & 49.2 & 79.6 & 82.5 & \underline{99.9} \\
    GDN & 71.4 & 80.6 & 33.4 & 59.0 & 87.1 & \textbf{99.9} & 76.4 & \textbf{100} & 40.8 & 66.2 & 86.1 & \textbf{100} & 85.1 & \textbf{100} & 38.7 & 56.7 & 93.8 & \textbf{100} \\
    MAD\_GAN & \underline{74.2} & 85.0 & \underline{51.5} & \underline{65.9} & \underline{99.4} & \textbf{99.9} & 61.3 & \textbf{100} & 39.9 & 72.3 & 83.3 & \textbf{100} & {96.0} & \textbf{100} & 46.4 & 50.0 & 95.7 & \textbf{100} \\
    MSCRED & 32.6 & 96.0 & 28.9 & 45.9 & 94.2 & \textbf{99.9} & 57.0 & \underline{97.9} & 40.8 & 61.7 & 77.0 & \textbf{100} & 63.0 & 92.2 & 39.5 & 51.8 & 73.2 & 98.1 \\
    MTAD\_GAT & 14.8 & 36.6 & 34.2 & 38.9 & 84.6 & {94.4} & 78.3 & \textbf{100} & 40.2 & 58.0 & 77.0 & \textbf{100} & 90.6 & \textbf{100} & 49.2 & 67.8 & 81.2 & \textbf{100} \\
    OmniAnomaly & 34.5 & 95.7 & 26.0 & 45.9 & 85.6 & \textbf{99.9} & 57.1 & \textbf{100} & 43.6 & 63.2 & 77.5 & \textbf{100} & 71.4 & \textbf{100} & 40.0 & 74.9 & 85.0 & \underline{99.9} \\
    USAD & 57.6 & \textbf{100} & 33.1 & 50.0 & 97.1 & \textbf{99.9} & 72.8 & \textbf{100} & 43.6 & 63.2 & 93.9 & \textbf{100} & 91.6 & \underline{99.9} & 42.6 & 60.8 & \underline{94.2} & \textbf{100} \\
    TimesNet & 32.8 & 45.8 & 15.4 & 23.5 & 98.4 & 99.4 & \textbf{97.7} & \textbf{100} & 51.4 & \underline{90.3} & \textbf{99.8} & \textbf{100} & \underline{97.4} & \textbf{100} & 52.9 & 79.7 & \textbf{99.8} & \textbf{100} \\
    SIGLLM~(GPT-4o) & 23.1 & 44.6 & 7.40 & 15.5 & 93.5 & \underline{96.5} & 69.0 & 97.8 & 29.1 & 49.2 & 95.5 & 99.8 & 70.7 & 97.9 & 72.4 & \textbf{100} & 90.0 & \textbf{100} \\
    \midrule
    \textbf{TAMA} & \textbf{92.5} & \underline{97.6} & \textbf{93.0} & \textbf{97.7} & \textbf{99.8} & \textbf{99.9} & \underline{94.5} & \textbf{100} & \textbf{95.5} & \textbf{100} & \underline{98.4} & \textbf{100} & \textbf{97.5} & \textbf{100} & \textbf{99.4} & \textbf{100} & \textbf{99.8} & \textbf{100} \\
    \textbf{TAMA*} & \textbf{92.5} & \underline{97.6} & \textbf{93.0} & \textbf{97.7} & \textbf{99.8} & \textbf{99.9} & {87.8} & \textbf{100} & \underline{89.2} & \textbf{100} & {97.0} & \textbf{100} & {96.1} & \textbf{100} & \underline{97.7} & \textbf{100} & {99.0} & \textbf{100} \\
    \midrule
    \midrule
    Dataset & \multicolumn{6}{c|}{\textbf{SMD}} & \multicolumn{6}{c|}{\textbf{ECG}} & \multicolumn{6}{c}{\textbf{Dodgers}}\\
    \midrule
    Metric& \multicolumn{2}{c|}{F1\%}& \multicolumn{2}{c|}{AUC-PR\%}& \multicolumn{2}{c|}{AUC-ROC\%}& \multicolumn{2}{c|}{F1\%}& \multicolumn{2}{c|}{AUC-PR\%}& \multicolumn{2}{c|}{AUC-ROC\%}& \multicolumn{2}{c|}{F1\%}& \multicolumn{2}{c|}{AUC-PR\%}& \multicolumn{2}{c}{AUC-ROC\%}\\
    \midrule
    IF & \textbf{83.9} & \textbf{100} & 73.8 & 97.0 & \underline{99.5} & \textbf{100} & {80.8} & 99.0 & 73.4 & 92.2 & \underline{97.2} & \textbf{100} & 48.4 & 48.4 & 52.2 & 52.2 & \textbf{89.4} & \textbf{89.4} \\
    LOF & 27.8 & 75.2 & 39.9 & 64.6 & 52.9 & 59.3 & 21.8 & 39.8 & 41.4 & 60.7 & 56.3 & 84.2 & 45.3 & 45.3 & 40.8 & 40.8 & 63.0 & 63.0 \\
    TranAD & 77.0 & 99.6 & 70.9 & 91.0 & 96.8 & \textbf{100} & 69.1 & 98.9 & 74.7 & 97.7 & 94.9 & \textbf{100} & 38.2 & 38.2 & 33.9 & 33.9 & 74.6 & 74.6 \\
    GDN & 76.9 & \underline{99.7} & 55.0 & 88.6 & 77.0 & \textbf{100} & 75.2 & 96.2 & 76.6 & 97.4 & {96.9} & \underline{99.9} & 37.0 & 37.0 & 31.3 & 31.3 & 74.2 & 74.2 \\
    MAD\_GAN & 67.1 & 92.6 & 61.7 & 87.5 & 91.7 & \textbf{100} & 79.1 & \underline{99.3} & 79.2 & 97.6 & {96.9} & \textbf{100} & 32.2 & 32.2 & 28.6 & 28.6 & 74.7 & 74.7 \\
    MSCRED & 69.4 & 95.7 & 55.8 & 96.2 & 95.0 & \textbf{100} & 66.4 & 99.0 & 73.8 & 97.2 & 88.7 & \textbf{100} & 37.8 & 37.8 & 30.6 & 30.6 & 74.6 & 74.6 \\
    MTAD\_GAT & 69.7 & 95.3 & 59.5 & 90.4 & 90.6 & \underline{99.7} & 67.5 & \textbf{100} & 73.5 & \textbf{98.8} & 82.5 & \textbf{100} & 39.1 & 39.1 & 36.0 & 36.0 & 74.9 & 74.9 \\
    OmniAnomaly & 66.0 & 96.4 & 61.6 & 91.8 & 87.7 & \textbf{100} & 76.8 & 98.6 & 76.4 & 97.5 & 93.5 & \textbf{100} & 33.6 & 33.6 & 35.4 & 35.4 & 60.3 & 60.3 \\
    USAD & 72.2 & \underline{99.7} & 67.8 & 93.5 & 94.4 & \textbf{100} & 71.5 & 96.9 & 75.2 & \underline{98.3} & 94.9 & \textbf{100} & 37.8 & 37.8 & 33.1 & 33.1 & 74.6 & 74.6 \\
    TimesNet & \underline{82.8} & \textbf{100} & 57.3 & \underline{99.9} & 95.4 & \textbf{100} & \textbf{92.4} & 96.6 & \textbf{90.0} & 97.6 & \textbf{99.4} & \textbf{100} & 48.1 & 48.1 & 73.0 & 73.0 & 83.7 & 83.7 \\
    SIGLLM~(GPT-4o) & 42.9 & 59.8 & 30.4 & 53.1 & 68.8 & 77.8 & 19.2 & 50.4 & 71.0 & 87.6 & 94.2 & 96.9 & 48.1 & 48.1 & 60.7 & 60.7 & 83.2 & 83.2 \\
    \midrule
    \textbf{TAMA} & {77.8} & \textbf{100} & \textbf{87.9} & \textbf{100} & 98.9 & \textbf{100} & \underline{81.3} & 87.5 & \underline{84.5} & 90.0 & 95.4 & 99.4 & \textbf{65.6} & \textbf{65.6} & \textbf{74.0} & \textbf{74.0} & 85.2 & 85.2 \\
    \textbf{TAMA*} & 62.8 & 93.0 & \underline{78.6} & {97.2} & \textbf{99.7} & \underline{99.7} & 78.1 & 88.0 & {83.4} & 91.1 & 94.7 & 99.1 & \underline{64.5} & \underline{64.5} & \underline{73.6} & \underline{73.6} & \underline{85.3} & \underline{85.3} \\
    \bottomrule
    \end{tabular}
    }
\end{table*}

%% file: tables/dataset.tex
\begin{table}[t]
\caption{Details of all datasets. Datasets with classification labels include real-world datasets we labeled (marked with $^{+}$) and a synthetic dataset (marked with $^{\ast}$) we generated using GutenTAG \cite{wenig_timeeval_2022}. "-" denotes corresponding training data does not exist or missing classification labels. }\label{tab:dataset}
    \vspace{-5pt}
    \setlength{\tabcolsep}{2pt}
    \centering
    \resizebox{\columnwidth}{!}{
    \begin{tabular}{c|c|c|ccccc}
      \toprule
      Dataset  & \#Train & \#Test (labeled) & \multicolumn{5}{c}{\textbf{Anomaly\%}} \\
       & & &Point&Shapelet&Seasonal&Trend&Total\\
      \midrule
       UCR$^{+}$  & 1,200-3,000 & 4,500-6,301 & 0.04&1.05&0&0&1.10 \\
       SMAP$^{+}$ &  312-2,881  & 4,453-8,640   &0&7.0&0.2&0.1&7.3 \\
       MSL$^{+}$  & 439-4,308 & 1,096-6,100  & 1.3&6.2&0&3.0&10.5 \\
       NormA$^{+}$ &   - & 104,000-196,000 & 0&18.6&4.1&1.2&24.0 \\
       Synthetic$^{\ast}$ &3,600&3,600&0.3&0&3.4&1.4&5.1\\
       SMD &23,687-28,743&23,687-28,743&-&-&-&-&4.2\\
       Dodgers &-&50,400&-&-&-&-&11.1\\
       ECG &227,900-267,228&227,900-267,228&-&-&-&-&7.9\\
       
      \bottomrule
    \end{tabular}
    }
\vspace{-15pt}
\end{table}

%% file: tables/classification.tex
\begin{table}[t]
\caption{Classification results are detailed for each anomaly type, with ‘total’ representing the overall performance.}
\vspace{-5pt}
\begin{tabular}{c|c|c|c|c|c}
    \toprule
    Type & \text{Point} & \text{Shapelet} & \text{Seasonal} & \text{Trend} & \textbf{Total} \\
    \midrule
    \text{Accuracy\%} & 81.0 & 99.2 & 29.0 & 74.5 & 78.5 \\
    \text{Support} & 100 & 246 & 100 & 94 & 567 \\
    \bottomrule
\end{tabular}
\label{tab:o_class_r}
\vspace{-12pt}
\end{table}

%% file: tables/ablation-LMM.tex
\begin{table}[b]
\vspace{-15pt}
\caption{
Comparison of different pre-trained LMMs using the average \emph{AUC-PR\%} without PA as the metric. 
We compare results with (\textit{+\textbf{TAMA}}) and without our framework (\textit{Naive}). 
}
\label{tab:ablation-LLMs}
    \vspace{-5pt}
    \centering
    \begin{tabular}{c|c|c}
      \toprule
      LMM  & \textit{Naive} & \textit{+\textbf{TAMA}} \\
      \midrule
        GPT-4o          & 41.8  &   80.2 (\textbf{+38.4})\\
        GPT-4o-mini     & 11.8  &   51.1 (\textbf{+39.3})\\
        Gemini-1.5-pro  & 25.4  &   87.8 (\textbf{+62.4})\\
        Gemini-1.5-flash& 17.9  &   36.4 (\textbf{+18.5})\\
        qwen-vl-max-0809& 61.7  &   80.5 (\textbf{+18.8})\\
      \bottomrule
    \end{tabular}
\vspace{-10pt}
\end{table}

%% file: tables/ablation-modality.tex
\begin{table}[thbp]
\vspace{-5pt}
\caption{
The performance comparison of image and text modalities, using the average PA \emph{F1\%} as the metric.
TAMA~(Image) and TAMA~(Text) are based on TAMA but using image-modality and text-modality respectively. 
}
\label{tab:ablation-modality}
    \vspace{-5pt}
    \centering
    \resizebox{0.7\columnwidth}{!}{
    \begin{tabular}{c|c|c}
      \toprule
      Modality  & NASA-MSL& NASA-SMAP\\
      \midrule
        TAMA~(Image)   & 97.5 & 94.5\\
        TAMA~(Text)    & 70.7 & 69.0\\
        SIGLLM~(Text)~\cite{alnegheimish_large_2024}   & 42.9 & 43.1\\
      \bottomrule
    \end{tabular}}
\vspace{-10pt}
\end{table}

%% file: tables/ablation-rotation.tex
\begin{table}[thbp]
\vspace{-5pt}
\caption{
The average \emph{AUC-PR}\% performance of TAMA with different additional information in images. 
}
\label{tab:ablation-rotation}
    \vspace{-5pt}
    \centering
    \resizebox{0.7\columnwidth}{!}{
    \begin{tabular}{c|c|c|c}
      \toprule
      Datasets  & TAMA& TAMA-R & TAMA-A\\
      \midrule
        UCR   & 83.0 & 32.9~(\textbf{-50.1}) & 75.6~(\textbf{-7.60})\\
        NASA-SMAP    & 72.9 & 28.6~(\textbf{-44.3})& 66.4~(\textbf{-6.50})\\
      \bottomrule
    \end{tabular}}
\end{table}

%% file: tables/ablation-NormalReference.tex
\begin{table}[thbp]
\vspace{-15pt}
\caption{
The performance comparison in different reference images, using the \emph{AUC-PR\%} without PA as the metric.
}
\label{tab:ablation-NR}
    \vspace{-5pt}
    \centering
    \resizebox{0.5\columnwidth}{!}{
    \begin{tabular}{c|c|c}
      \toprule
      Dataset  & \textit{normal} & \textit{abnormal}\\
      \midrule
        UCR  &   83.0   &  46.8~(\textbf{-36.2}) \\
        NASA-SMAP&   72.9   &   48.5~(\textbf{-24.4})\\
      \bottomrule
    \end{tabular}}
\vspace{-10pt}
\end{table}

%% file: tables/full_results.tex
\begin{table*}[t!]
    \caption{Quantitative results across six datasets use metrics point-adjusted \emph{F1}, \emph{AUC-PR}, and \emph{AUC-ROC}. Best and second-best results are in bold and underlined, respectively. \textbf{TAMA} represents our framework, and \textbf{TAMA*} represents our framework without self-reflection. Each unit in the table contains two value: \emph{mean} and \emph{maxima} of all series. The number following the mean represents the standard deviation (\emph{std}) computed over all sequences.}
    \label{tab:appendix_main_results}
    \centering
    \resizebox{\textwidth}{!}{
    \renewcommand{\multirowsetup}{\centering}
    \begin{tabular}{c|cccccc|cccccc|cccccc}
    \toprule
    Dataset & \multicolumn{6}{c|}{\textbf{UCR}} & \multicolumn{6}{c|}{\textbf{NASA-SMAP}}& \multicolumn{6}{c}{\textbf{NASA-MSL}}\\
    \midrule
    Metric& \multicolumn{2}{c|}{F1\%}& \multicolumn{2}{c|}{AUC-PR\%}& \multicolumn{2}{c|}{AUC-ROC\%}& \multicolumn{2}{c|}{F1\%}& \multicolumn{2}{c|}{AUC-PR\%}& \multicolumn{2}{c|}{AUC-ROC\%}& \multicolumn{2}{c|}{F1\%}& \multicolumn{2}{c|}{AUC-PR\%}& \multicolumn{2}{c}{AUC-ROC\%}\\
    \midrule
    IF  & 24.7 $\pm$ 31.6 & 77.3 & 37.7 $\pm$ 15.0 & 44.6 & 24.4 $\pm$ 9.80 & 25.3 & 54.2 $\pm$ 36.4 & 94.2 & 58.9 $\pm$ 18.9 & 77.1 & 65.0 $\pm$ 6.10 & 87.7 & 47.6 $\pm$ 8.00 & 88.6 & 53.6 $\pm$ 4.80 & \underline{80.4} & 68.7 $\pm$ 9.40 & 88.7 \\
    LOF & 42.8 $\pm$ 1.10 & \textbf{100} & 35.6 $\pm$ 1.00 & 50.0 & 92.8 $\pm$ 19.2 & \textbf{99.9} & 62.2 $\pm$ 12.1 & \textbf{100} & 43.4 $\pm$ 12.9 & 61.4 & 60.1 $\pm$ 19.9 & \underline{99.9} & 36.4 $\pm$ 25.3 & 66.8 & 44.5 $\pm$ 9.90 & 66.0 & 58.6 $\pm$ 0.50 & 99.8 \\
    TranAD & 38.2 $\pm$ 40.7 & 93.7 & 30.9 $\pm$ 6.40 & 51.0 & 77.0 $\pm$ 26.8 & \textbf{99.9} & 59.0 $\pm$ 39.9 & 99.6 & 36.8 $\pm$ 19.1 & 73.9 & 74.4 $\pm$ 27.7 & \textbf{100} & 64.6 $\pm$ 38.6 & 99.1 & 49.2 $\pm$ 20.4 & 79.6 & 82.5 $\pm$ 18.3 & \underline{99.9} \\
    GDN & 71.4 $\pm$ 43.1 & 80.6 & 33.4 $\pm$ 0.40 & 59.0 & 87.1 $\pm$ 24.1 & \textbf{99.9} & 76.4 $\pm$ 38.5 & \textbf{100} & 40.8 $\pm$ 19.1 & 66.2 & 86.1 $\pm$ 27.7 & \textbf{100} & 85.1 $\pm$ 26.1 & \textbf{100} & 38.7 $\pm$ 9.90 & 56.7 & 93.8 $\pm$ 0.50 & \textbf{100} \\
    MAD\_GAN & \underline{74.2} $\pm$ 40.4 & 85.0 & \underline{51.5} $\pm$ 0.80 & \underline{65.9} & \underline{99.4} $\pm$ 1.30 & \textbf{99.9} & 61.3 $\pm$ 41.3 & \textbf{100} & 39.9 $\pm$ 19.1 & 72.3 & 83.3 $\pm$ 15.1 & \textbf{100} & {96.0} $\pm$ 5.60 & \textbf{100} & 46.4 $\pm$ 17.5 & 50.0 & 95.7 $\pm$ 7.10 & \textbf{100} \\
    MSCRED & 32.6 $\pm$ 37.9 & 96.0 & 28.9 $\pm$ 2.70 & 45.9 & 94.2 $\pm$ 3.60 & \textbf{99.9} & 57.0 $\pm$ 44.2 & \underline{97.9} & 40.8 $\pm$ 19.1 & 61.7 & 77.0 $\pm$ 28.2 & \textbf{100} & 63.0 $\pm$ 37.0 & 92.2 & 39.5 $\pm$ 16.3 & 51.8 & 73.2 $\pm$ 16.1 & 98.1 \\
    MTAD\_GAT & 14.8 $\pm$ 13.2 & 36.6 & 34.2 $\pm$ 4.80 & 38.9 & 84.6 $\pm$ 7.60 & \underline{94.4} & 78.3 $\pm$ 37.7 & \textbf{100} & 40.2 $\pm$ 22.9 & 58.0 & 77.0 $\pm$ 12.3 & \textbf{100} & 90.6 $\pm$ 27.7 & \textbf{100} & 49.2 $\pm$ 9.70 & 67.8 & 81.2 $\pm$ 21.9 & \textbf{100} \\
    OmniAnomaly & 34.5 $\pm$ 32.7 & 95.7 & 26.0 $\pm$ 0.30 & 45.9 & 85.6 $\pm$ 10.3 & \textbf{99.9} & 57.1 $\pm$ 39.9 & \textbf{100} & 43.6 $\pm$ 20.5 & 63.2 & 77.5 $\pm$ 24.8 & \textbf{100} & 71.4 $\pm$ 36.5 & \textbf{100} & 40.0 $\pm$ 17.2 & 74.9 & 85.0 $\pm$ 18.7 & \underline{99.9} \\
    USAD & 57.6 $\pm$ 35.6 & \textbf{100} & 33.1 $\pm$0.40 & 50.0 & 97.1 $\pm$ 4.20 & \textbf{99.9} & 72.8 $\pm$ 35.8 & \textbf{100} & 43.6 $\pm$ 22.5 & 63.2 & 93.9 $\pm$ 9.10 & \textbf{100} & 91.6 $\pm$ 26.2 & \underline{99.9} & 42.6 $\pm$ 9.90 & 60.8 & {94.2} $\pm$ 0.70 & \textbf{100} \\
    TimesNet & 32.8 $\pm$ 8.30 & 45.8 & 15.4 $\pm$ 5.20 & 23.5 & 98.4 $\pm$ 1.10 & 99.4 & \textbf{97.7} $\pm$ 3.50 & \textbf{100} & 51.4 $\pm$ 2.80 & \underline{90.3} & \textbf{99.8} $\pm$ 0.09 & \textbf{100} & \underline{97.4} $\pm$ 4.70 & \textbf{100} & 52.9 $\pm$ 8.10 & 79.7 & \textbf{99.8} $\pm$ 0.50 & \textbf{100} \\
    SIGLLM~(GPT-4o) & 23.1 $\pm$ 19.7 & 44.6 & 7.40 $\pm$ 6.70 & 15.5 & 93.5 $\pm$ 16.9 & 96.5 & 69.0 $\pm$ 34.4 & 97.8 & 29.1 $\pm$ 28.4 & 49.2 & 95.5 $\pm$ 3.60 & 99.8 & 70.7 $\pm$ 44.8 & 97.9 & 72.4 $\pm$ 28.6 & \textbf{100} & 90.0 $\pm$ 15.3 & \textbf{100} \\
    \midrule
     \textbf{TAMA} & \textbf{92.5} $\pm$ 17.9 & \underline{97.6} & \textbf{93.0} $\pm$ 12.1 & \textbf{97.7} & \textbf{99.8} $\pm$ 0.10 & \textbf{99.9} & \underline{94.5} $\pm$ 7.20 & \textbf{100} & \textbf{95.5} $\pm$ 9.30 & \textbf{100} & \underline{98.4} $\pm$ 4.60 & \textbf{100} & \textbf{97.5} $\pm$ 2.10 & \textbf{100} & \textbf{99.4} $\pm$ 17.8 & \textbf{100} & \textbf{99.8} $\pm$ 0.20 & \textbf{100} \\
    \textbf{TAMA*} & \textbf{92.5} $\pm$ 17.9 & \underline{97.6} & \textbf{93.0} $\pm$ 12.1 & \textbf{97.7} & \textbf{99.8} $\pm$ 0.10 & \textbf{99.9} & {87.8} $\pm$ 31.3 & \textbf{100} & \underline{89.2} $\pm$ 16.6 & \textbf{100} & {97.0} $\pm$ 4.10 & \textbf{100} & {96.1}$\pm$ 4.30 & \textbf{100} & \underline{97.7} $\pm$ 18.2 & \textbf{100} & \underline{99.0} $\pm$ 0.20 & \textbf{100} \\
    \midrule
    \midrule
    Dataset & \multicolumn{6}{c|}{\textbf{SMD}} & \multicolumn{6}{c|}{\textbf{ECG}} & \multicolumn{6}{c}{\textbf{Dodgers}}\\
    \midrule
    Metric& \multicolumn{2}{c|}{F1\%}& \multicolumn{2}{c|}{AUC-PR\%}& \multicolumn{2}{c|}{AUC-ROC\%}& \multicolumn{2}{c|}{F1\%}& \multicolumn{2}{c|}{AUC-PR\%}& \multicolumn{2}{c|}{AUC-ROC\%}& \multicolumn{2}{c|}{F1\%}& \multicolumn{2}{c|}{AUC-PR\%}& \multicolumn{2}{c}{AUC-ROC\%}\\
    \midrule
    IF & \textbf{83.9} $\pm$ 13.2 & \textbf{100} & 73.8 $\pm$ 17.3 & 97.0 & \underline{99.5} $\pm$ 0.50 & \textbf{100} & {80.8} $\pm$ 20.5 & 99.0 & 73.4 $\pm$ 18.8 & 92.2 & \underline{97.2} $\pm$ 4.70 & \textbf{100} & 48.4 $\pm$ 0.00 & 48.4 & 52.2 $\pm$ 0.00 & 52.2 & \textbf{89.4} $\pm$ 0.00 & \textbf{89.4} \\
    LOF & 27.8 $\pm$ 6.60 & 75.2 & 39.9 $\pm$ 1.90 & 64.6 & 52.9 $\pm$ 2.60 & 59.3 & 21.8 $\pm$ 12.0 & 39.8 & 41.4 $\pm$ 4.40 & 60.7 & 56.3 $\pm$ 10.8 & 84.2 & 45.3 $\pm$ 0.00 & 45.3 & 40.8 $\pm$ 0.00 & 40.8 & 63.0 $\pm$ 0.00 & 63.0 \\
    TranAD & 77.0 $\pm$ 33.0 & 99.6 & 70.9 $\pm$ 31.9 & 91.0 & 96.8 $\pm$ 13.3 & \textbf{100} & 69.1 $\pm$ 23.2 & 98.9 & 74.7 $\pm$ 22.9 & 97.7 & 94.9 $\pm$ 6.30 & \textbf{100} & 38.2 $\pm$ 0.00 & 38.2 & 33.9 $\pm$ 0.00 & 33.9 & 74.6 $\pm$ 0.00 & 74.6 \\
    GDN & 76.9 $\pm$ 0.90 & \underline{99.7} & 55.0 $\pm$ 35.5 & 88.6 & 77.0 $\pm$ 1.60 & \textbf{100} & 75.2 $\pm$ 17.6 & 96.2 & 76.6 $\pm$ 17.5 & 97.4 & {96.9} $\pm$ 3.70 & \underline{99.9} & 37.0 $\pm$ 0.00 & 37.0 & 31.3 $\pm$ 0.00 & 31.3 & 74.2 $\pm$ 0.00 & 74.2 \\
    MAD\_GAN & 67.1 $\pm$ 1.30 & 92.6 & 61.7 $\pm$ 37.2 & 87.5 & 91.7 $\pm$ 1.60 & \textbf{100} & 79.1 $\pm$ 19.5 & \underline{99.3} & 79.2 $\pm$ 19.4 & 97.6 & {96.9} $\pm$ 3.80 & \textbf{100} & 32.2 $\pm$ 0.00 & 32.2 & 28.6 $\pm$ 0.00 & 28.6 & 74.7 $\pm$ 0.00 & 74.7 \\
    MSCRED & 69.4 $\pm$ 30.2 & 95.7 & 55.8 $\pm$ 36.7 & 96.2 & 95.0 $\pm$ 14.5 & \textbf{100} & 66.4 $\pm$ 26.3 & 99.0 & 73.8 $\pm$ 22.6 & 97.2 & 88.7 $\pm$ 13.5 & \textbf{100} & 37.8 $\pm$ 0.00 & 37.8 & 30.6 $\pm$ 0.00 & 30.6 & 74.6 $\pm$ 0.00 & 74.6 \\
    MTAD\_GAT & 69.7 $\pm$ 18.5 & 95.3 & 59.5 $\pm$ 33.8  & 90.4 & 90.6 $\pm$ 6.00 & \underline{99.7} & 67.5 $\pm$ 29.2 & \textbf{100} & 73.5 $\pm$ 23.8 & \textbf{98.8} & 82.5 $\pm$ 17.1 & \textbf{100} & 39.1 $\pm$ 0.00 & 39.1 & 36.0 $\pm$ 0.00 & 36.0 & 74.9 $\pm$ 0.00 & 74.9 \\
    OmniAnomaly & 66.0 $\pm$ 6.80 & 96.4 & 61.6 $\pm$ 34.2 & 91.8 & 87.7 $\pm$ 2.60 & \textbf{100} & 76.8 $\pm$ 21.3 & 98.6 & 76.4 $\pm$ 24.0 & 97.5 & 93.5 $\pm$ 8.80 & \textbf{100} & 33.6 $\pm$ 0.00 & 33.6 & 35.4 $\pm$ 0.00 & 35.4 & 60.3 $\pm$ 0.00 & 60.3 \\
    USAD & 72.2 $\pm$ 0.30 & \underline{99.7} & 67.8 $\pm$ 35.6 & 93.5 & 94.4 $\pm$ 1.60 & \textbf{100} & 71.5 $\pm$ 20.8 & 96.9 & 75.2 $\pm$ 18.9 & \underline{98.3} & 94.9 $\pm$ 5.90 & \textbf{100} & 37.8 $\pm$ 0.00 & 37.8 & 33.1 $\pm$ 0.00 & 33.1 & 74.6 $\pm$ 0.00 & 74.6 \\
    TimesNet & \underline{82.8} $\pm$ 25.3 & \textbf{100} & 57.3 $\pm$ 21.2 & \underline{99.9} & 95.4 $\pm$ 11.3 & \textbf{100} & \textbf{92.4} $\pm$ 3.70 & 96.6 & \textbf{90.0} $\pm$ 5.60 & 97.6 & \textbf{99.4} $\pm$ 0.40 & \textbf{100} & 48.1 $\pm$ 0.00 & 48.1 & 73.0 $\pm$ 0.00 & 73.0 & 83.7 $\pm$ 0.00 & 83.7 \\
    SIGLLM~(GPT-4o) & 42.9 $\pm$ 27.9 & 59.8 & 30.4 $\pm$ 21.0 & 53.1 & 68.8 $\pm$ 12.8 & 77.8 & 19.2 $\pm$ 13.7 & 50.4 & 71.0 $\pm$ 25.3 & 87.6 & 94.2 $\pm$ 3.20 & 96.9 & 48.1 $\pm$ 0.00 & 48.1 & 60.7 $\pm$ 0.00 & 60.7 & 83.2 $\pm$ 0.00 & 83.2 \\
    \midrule
    \textbf{TAMA} & {77.8} $\pm$ 17.1 & \textbf{100} & \textbf{87.9} $\pm$ 10.4 & \textbf{100} & 98.9 $\pm$ 1.40 & \textbf{100} & \underline{81.3} $\pm$ 19.1 & 87.5 & \underline{84.5} $\pm$ 15.4 & 90.0 & 95.4 $\pm$ 2.30 & 99.4 & \textbf{65.6} $\pm$ 0.00 & \textbf{65.6} & \textbf{74.0} $\pm$ 0.00 & \textbf{74.0} & 85.2 $\pm$ 0.00 & 85.2 \\
    \textbf{TAMA*} & 62.8 $\pm$ 24.5 & 93.0 & \underline{78.6} $\pm$ 14.1 & {97.2} & \textbf{99.7} $\pm$ 1.50 & \underline{99.7} & 78.1 $\pm$ 19.8 & 88.0 & {83.4} $\pm$ 14.6 & 91.1 & 94.7 $\pm$ 2.50 & 99.1 & \underline{64.5} $\pm$ 0.00 & \underline{64.5} & \underline{73.6} $\pm$ 0.00 & \underline{73.6} & \underline{85.3} $\pm$ 0.00 & \underline{85.3} \\
    \midrule
    \midrule
    Dataset & \multicolumn{6}{c|}{\textbf{NormA}}\\
    \midrule
    Metric& \multicolumn{2}{c|}{F1\%}& \multicolumn{2}{c|}{AUC-PR\%}& \multicolumn{2}{c|}{AUC-ROC\%}\\
    \midrule
    IF & 56.8 $\pm$ 19.2 & 86.3 & 52.3 $\pm$ 21.9 & 81.2 & 57.9 $\pm$ 1.00 & 68.7 \\
    LOF & 54.5 $\pm$ 17.8 & 77.9 & 68.8 $\pm$ 9.30 & 92.4 & 95.1 $\pm$ 2.90 & 97.9 \\
    TranAD & 38.0 $\pm$ 15.8 & 76.0 & 49.7 $\pm$ 21.3 & 78.9 & 53.6 $\pm$ 2.00 & 83.6 \\
    GDN & 38.5 $\pm$ 14.8 & 74.7 & 50.9 $\pm$ 20.2 & 78.3 & 54.2 $\pm$ 2.10 & 82.2 \\
    MAD\_GAN & 38.5 $\pm$ 14.3 & 74.7 & 51.1 $\pm$ 19.8 & 77.8 & 54.1 $\pm$ 2.90 & 81.9 \\
    MSCRED & 38.4 $\pm$ 16.1 & 74.6 & 49.7 $\pm$ 20.9 & 77.7 & 53.8 $\pm$ 2.00 & 81.8 \\
    MTAD\_GAT & 49.7 $\pm$ 13.7 & \underline{93.8} & 50.1 $\pm$ 21.3 & 95.6 & 66.6 $\pm$ 3.30 & 94.2 \\
    OminiAnomaly & 43.2 $\pm$ 17.9 & 74.8 & 53.5 $\pm$ 20.3 & 79.1 & 49.8 $\pm$ 1.70 & 89.9 \\
    USAD & 38.6 $\pm$ 15.9 & 75.6 & 53.3 $\pm$ 20.9 & 78.5 & 54.1 $\pm$ 1.70 & 82.9 \\
    SIGLLM~(GPT-4o) & \underline{82.8} $\pm$ 30.0 & \textbf{94.6} & 93.8 $\pm$ 21.4 & \textbf{98.9} & \underline{97.9} $\pm$ 2.50 & \underline{99.1} \\
    \midrule
    \textbf{TAMA} & {80.7} $\pm$ 4.70 & {89.2} & \textbf{95.0} $\pm$ 7.60 & {98.5} & \textbf{98.1} $\pm$ 0.70 & \textbf{99.2} \\
    \textbf{TAMA*} & \textbf{83.9} $\pm$ 10.0 & 85.5 & \underline{93.9} $\pm$ 10.8 & \underline{98.7} & 97.4 $\pm$ 1.00 & {98.6} \\
    \bottomrule
    \end{tabular}
    }
\end{table*}

%% file: tables/type-wise-eval.tex
\begin{table*}[t!]
    \caption{Quantitative results on each specific anomaly category across five datasets using F1-score\% without point-adjustment. Best and second-best results are in bold and underlined, respectively. \textbf{TAMA} represents our framework  (Some datasets include more than one series. To present the true performance of each method as much as possible, each unit in the table contains two values: \emph{maxima} / \emph{mean}. The \emph{maxima} represents the best result among all sub-series, while the \emph{mean} refers to the average of all sub-series.).}
    \label{tab:appendix_wise_eval}
    \centering
    \resizebox{0.9\textwidth}{!}{
    \renewcommand{\multirowsetup}{\centering}
    \begin{tabular}{c|ccc|ccc|cc}
    \toprule
    Dataset & \multicolumn{3}{c|}{\textbf{NASA-MSL}}& \multicolumn{3}{c|}{\textbf{NASA-SMAP}}& \multicolumn{2}{c}{\textbf{UCR}} \\
    Category&  Point&  Shapelet& Trend& Shapelet& Seasonal& Trend& Point& Shapelet\\
    \midrule
     TranAD      & 23.2 / 10.4    & 14.9 / 13.5  & \underline{33.7} / \underline{33.7}  & 1.00 / 0.60       & 1.40 / 1.40       & 0.30 / 0.30       & 0.30 / 0.30   & 10.5 / 3.60  \\
     GDN         & 3.80 / 2.30      & 17.6 / 8.60     & 1.20 / 1.20      & 6.30 / 2.10       & 1.20 / 1.20       & 0.30 / 0.30       & \textbf{22.2} / \textbf{22.2} & \underline{53.0} / \underline{20.6} \\
     MAD\_GAN     & 3.90 / 2.00      & 17.6 / 8.50     & 1.60 / 1.60      & 10.2 / 3.00      & 1.20 / 1.20       & 0.65 / 0.65     & 14.0 / 14.0 & 36.8 / 15.0 \\
     MSCRED      & 61.9 / 23.0    & 16.6 / 8.00     & 23.4 / 23.4    & 1.00 / 0.60      & 0.70 / 0.70       & 0.30 / 0.30       & 0.45 / 0.45 & 2.40 / 0.80   \\
     MTAD\_GAT    & \underline{69.8} / \textbf{46.3}    & 16.6 / 8.00     & \textbf{73.9} / \textbf{73.9}    & \underline{52.0} / \underline{24.0}     & \underline{2.10} / \underline{2.10}       & 0.95 / 0.95     & 0.80 / 0.80   & 2.70 / 1.45  \\
     OmniAnomaly & 5.40 / 1.80      & 3.70 / 1.80      & 2.80 / 2.80      & 1.00 / 0.40       & 0.65 / 0.65     & 0.45 / 0.45     & 0.55 / 0.55 & 3.00 / 1.00   \\
     USAD        & 13.0 / 7.50     & 16.6 / 8.10     & 4.30 / 4.30       & 0.60 / 0.30       & 1.20 / 1.20       & 1.25 / 1.25     & 5.30 / 5.30   & 11.9 / 4.50  \\
     IF          & 35.0 / 24.2   & \underline{30.0} / \textbf{22.9}    & 31.6 / 31.6  & 30.2 / 17.1     & 1.10 / 1.10       & 1.45 / 1.45     & 0.55 / 0.55 & 2.70 / 1.85  \\
     LoF         & 22.9 / 10.1  & \textbf{33.4} / \underline{22.4}  & 33.3 / 33.3    & 14.0 / 7.80      & 0.60 / 0.60       & 4.60 / 4.60       & 0.45 / 0.45 & 2.05 / 1.35 \\
     TimesNet    & 23.2 / 10.9   & 12.5 / 8.30    & 22.4 / 10.8   & 20.4 / 8.95     & 1.35 / 1.35     & 25.6 / 25.6   & 0.40 / 0.40   & 1.80 / 1.20   \\
     SIGLLM & 10.8 / 5.70     & 1.60 / 0.80      & 23.9 / 23.9  & 30.3 / 12.6     & \textbf{20.2} / \textbf{20.2}    & 2.65 / 2.65     & 0.85 / 0.85 & 11.2 / 4.80   \\
     \midrule
     TAMA        & \textbf{70.2} / \underline{31.9}    & 26.2 / 11.4    & 22.4 / 13.5    & \textbf{77.4} / \textbf{47.9}     & 0.10 / 0.10       & \textbf{84.5} / \textbf{84.5}     & \underline{20.0} / \underline{20.0} & \textbf{92.3} / \textbf{81.0} \\

    \midrule
    \midrule
    Dataset & \multicolumn{3}{c|}{\textbf{NormA}}& \multicolumn{3}{c|}{\textbf{Synthetic}} \\
    Category& Shapelet& Seasonal& Trend& Point& Seasonal& Trend\\
    \midrule
     TranAD      & 4.00 / 2.30    & 3.30 / 2.20   & 3.90 / 2.50    & 0.55 / 0.35              & 8.90 / 1.90                & 13.5 / \underline{7.90}               \\
     GDN         & 4.10 / 2.30    & 3.30 / 2.20   & 3.90 / 2.50    & 0.50 / 0.35               & 13.6 / 2.10               & 12.4 / 6.50               \\
     MAD\_GAN     & 4.10 / 2.30    & 3.30 / 2.20   & 3.90 / 2.50    & 0.55 / 0.35              & 8.50 / 1.60                & 13.9 / 8.00               \\
     MSCRED      & 4.10 / 2.30   & 3.30 / 2.20   & 3.90 / 2.50    & 0.55 / 0.35              & 8.50 / 1.60                & 10.5 / 5.20              \\
     MTAD\_GAT    & 4.90 / 1.40    & 1.30 / 0.90   & 1.70 / 1.10   & 0.55 / 0.40              & 6.30 / 1.80                & 13.3 / 6.25               \\
     OmniAnomaly & 12.6 / 5.70   & 11.5 / 7.50  & 11.4 / 7.30   & 30.3 / 19.8              & 9.30 / 1.70                & 11.2 / 7.80               \\
     USAD        & 4.10 / 2.40    & 3.30 / 2.20   & 3.90 / 2.50    & 0.55 / 0.35              & 8.90 / 1.70                & 12.6 / 7.30               \\
     IF          & 21.4 / 13.2  & 17.0 / 12.9 & 13.9 / 11.9 & \underline{36.2} / \underline{21.6}              & 10.4 / 9.05              & 10.9 / 7.20              \\
     LoF         & \underline{30.7} / \underline{16.8} & \underline{25.7} / \underline{18.6} & \underline{21.5} / \underline{16.9} & 0.50 / 0.50               & 10.9 / 9.10              & 5.35 / 5.25              \\
     TimesNet    & 10.2 / 9.05  & 5.25 / 5.20  & 1.79 / 1.60    & \textbf{37.5} / \textbf{25.5}              & 11.9 / \underline{9.50}              & 10.6 / 5.70             \\
     SIGLLM & 6.50 / 3.10   & 3.70 / 2.50   & 0.90 / 0.60  & 0.60 / 0.35               & \underline{10.9} / 7.95             & \underline{14.0} / 6.50               \\
     \midrule
     TAMA        & \textbf{56.8} / \textbf{37.1}  & \textbf{38.8} / \textbf{28.1} & \textbf{45.2} / \textbf{34.3}  & 3.90 / 1.80               & \textbf{27.1} / \textbf{18.4}             & \textbf{14.1} / \textbf{8.20}              \\

    \bottomrule
    \end{tabular}
    }
\end{table*}